%% file: main.tex
\documentclass[lettersize,journal]{IEEEtran}
\usepackage{amsmath,amsfonts}
\usepackage{array}
\usepackage[caption=false,font=normalsize,labelfont=sf,textfont=sf]{subfig}
\usepackage{textcomp}
\usepackage{stfloats}
\usepackage{url}
\usepackage{verbatim}
\usepackage{graphicx}
\usepackage{cite}
\usepackage{algorithm}
\usepackage{algorithmicx}
\usepackage{algpseudocode}
\usepackage{booktabs}
\usepackage{xcolor}
\usepackage{soul, color, xcolor}
\usepackage{threeparttable}

\soulregister{\cite}7 
\soulregister{\citep}7 
\soulregister{\citet}7 
\soulregister{\ref}7 
\soulregister{\pageref}7 

\begin{document}

\title{Hybrid Action Based Reinforcement Learning for Multi-Objective Compatible Autonomous Driving}


\author{Guizhe Jin, Zhuoren Li, Bo Leng, Wei Han, Lu Xiong, and Chen Sun  
\thanks{Received 22 January 2025; revised 20 August 2025; revised 17 December 2025; accepted 12 March 2026. This work is supported in part by the National Natural Science Foundation of China under Grant 52522219, under Grant 52372317 and under Grant 52232015, in part by Shanghai Automotive Industry Science and Technology Development Foundation under Grant 2203,  in part by the Fundamental Research Funds for the Central Universities under Grant 22120230311, in part by SAIC Motor Corporation Limited under Grant 2023023. \textit{(Corresponding author: Bo Leng.)}}
\thanks{Guizhe Jin, Zhuoren Li, Bo Leng, Wei Han and Lu Xiong are with the College of Automotive and Energy Engineering, Tongji University, Shanghai 201804, China. (Email: jgz13573016892@163.com, 1911055@tongji.edu.cn, lengbo@tongji.edu.cn, tjhanwei@foxmail.com, xiong\_lu@tongji.edu.cn).

Chen Sun is with the Department of Data and Systems Engineering, University of Hong Kong, Hong Kong. (Email: c87sun@hku.hk).

This work has been submitted to the IEEE for possible publication. Copyright may be transferred without notice, after which this version may no longer be accessible.

}
}

\markboth{IEEE TRANSACTIONS ON NEURAL NETWORKS AND LEARNING SYSTEMS}%
{Shell \MakeLowercase{\textit{et al.}}: A Sample Article Using IEEEtran.cls for IEEE Journals}

\maketitle

\begin{abstract}
Reinforcement Learning (RL) has shown excellent performance in solving decision-making and control problems of autonomous driving, which is increasingly applied in diverse driving scenarios. However, driving is a multi-attribute problem, leading to challenges in achieving multi-objective compatibility for current RL methods, especially in both policy updating and policy execution. On the one hand, a single value evaluation network limits the policy updating in complex scenarios with coupled driving objectives. On the other hand, the common single-type action space structure limits driving flexibility or results in large behavior fluctuations during policy execution. To this end, we propose a Multi-objective Ensemble-Critic reinforcement learning method with Hybrid Parametrized Action for multi-objective compatible autonomous driving. Specifically, an advanced MORL architecture is constructed, in which the ensemble-critic focuses on different objectives through independent reward functions. The architecture integrates a hybrid parameterized action space structure, and the generated driving actions contain both abstract guidance that matches the hybrid road modality and concrete control commands. Additionally, an uncertainty-based exploration mechanism that supports hybrid actions is developed to learn multi-objective compatible policies more quickly. Experimental results demonstrate that, in both simulator-based and HighD dataset-based multi-lane highway scenarios, our method efficiently learns multi-objective compatible autonomous driving with respect to efficiency, action consistency, and safety.
\end{abstract}


\input{I_Introduction}

\input{II_Related_Works}

\input{III_Methodology}

\input{IV_Implementation}

\input{V_Results_and_Discussions}

\input{VI_Conclusion}


\bibliographystyle{ieeetr}
\bibliography{reference}

\end{document}

%% file: I_Introduction.tex
\section{Introduction} \label{sec:Introduction}

Reinforcement learning (RL) has good potential in solving temporal decision-making problems~\cite{2023TNNLS_Majid}, which can learn viable and near-optimal policies for complex tasks~\cite{2021TNNLS_Zhang}. The RL agent explores policies through interactions with the environment, enabling self-improvement~\cite{2021TNNLS_Hao, 2022TNNLS_He}. The autonomous driving (AD) problem constitutes a complex sequential decision-making challenge~\cite{2021TM_Gao, 2021SCIS_Gao}. Therefore, RL is considered as an effective way to solve decision-making and control problems for AD~\cite{2021TNNLS_Xing}, particularly deep reinforcement learning (DRL) incorporating neural network techniques\cite{2025RAL_Jin}. It has led to widespread application in driving scenarios~\cite{OurSAE} and has outperformed human drivers in certain tasks~\cite{2022Nature_Wurman}. 


However, current RL methods still face several limitations in achieving compatibility with key driving objectives such as safety, efficiency, and action consistency~\cite{2023AI_knox, 2022TITS_Zhu}. In particular, when addressing multi-attribute driving tasks, mainstream RL-based AD approaches exhibit shortcomings in both policy update and execution: (i) For policy updates, most rely on a single critic (value network) to evaluate and guide learning, making it difficult to efficiently explore multi-objective-compatible policies within a large and complex traffic state space; (ii) For policy execution, most approaches employ a single-type action space structure to handle hybrid road modality, which limits the policy’s ability to fully represent real driving behaviors and forces a trade-off among certain objectives.

In terms of policy update, employing a single critic (i.e., a single reward function) to evaluate policy performance fails to capture the strong coupling and potential conflicts among driving objectives. When multiple attributes of an AD task are combined into a single reward function, the agent may disproportionately focus on certain attributes during training~\cite{2023AI_knox}. As a result, some objectives may be neglected in specific states, leading to inaccurate value estimation and suboptimal policy performance. This can cause the agent to behave in ways that are misaligned with multi-objective expectations, such as becoming overly aggressive to maximize speed or excessively conservative to ensure safety. In contrast, multi-objective reinforcement learning (MORL) addresses this issue more effectively by constructing reward function vectors \cite{2024NIPS_Cai}, enabling better compatibility among multiple objectives. Furthermore, complex traffic state spaces demand efficient exploration during policy updates. Most existing RL-based AD methods rely on random exploration, which prevents the agent from actively seeking out unknown regions and discovering potentially effective policies~\cite{2022IF_Ladosz}. This random exploration often results in the collection of redundant experiences that contribute little to policy improvement, leading to inefficient convergence or even entrapment in local optima.

For policy execution, using a single-type action space to generate either abstract or concrete driving behaviors constrains RL agents to discrete actions that lack flexibility or continuous actions that lack consistency. A common approach is to have the agent produce discrete, long-term driving goals, such as semantic decisions~\cite{2023TIV_Li} or target points for path planning \cite{ArXiv2023_lu}. However, because the agent does not directly control the vehicle’s motion, its ability to adapt driving behavior flexibly is limited. While this long-term planning enhances action consistency, it reduces responsiveness to dynamic changes. Conversely, directly outputting short-term control commands \cite{2022TVT_Chen} allows for greater flexibility, but often results in less consistent behavior, with frequent fluctuations and abrupt reactions to environmental changes.


To alleviate the limitations of policy updating and execution in typical multi-objective AD tasks, this paper proposes a \textbf{M}ulti-\textbf{o}bjective \textbf{E}nsemble-\textbf{C}ritic reinforcement learning method with \textbf{H}ybrid \textbf{P}arametrized \textbf{A}ction space (\textbf{HPA-MoEC}) for multi-objective compatibility. The HPA-MoEC adopts a MORL architecture focused on AD tasks. By defining multiple reward functions to decouple different driving attributes, each reward function guides an ensemble-critic to focus on specific driving objectives, thereby assisting the actor (policy network) in learning multi-objective compatible driving behaviors. The architecture further integrates a hybrid parameterized action space structure containing a discrete action set and its corresponding continuous parameters, which together generate driving actions that combine abstract guidance and concrete control commands. Additionally, uncertainty estimates from the ensemble-critic guide the agent to explore promising driving policies, facilitating more efficient exploration in unknown environments. Evaluation results on both simulation and HighD dataset-based multi-lane highway scenarios demonstrate that HPA-MoEC efficiently learns multi-objective compatible driving behaviors, significantly improving driving efficiency, action consistency, and safety. The main contributions are summarized as follows:

\begin{enumerate}
\item{A MORL architecture compatible with multiple AD objectives is proposed, in which ensemble-critic focuses on a distinct objective using separate reward functions. Considering the safety-critical nature of AD, two driving objectives are defined and evaluated with two ensemble critics: one targeting overall performance, including interactivity, and the other dedicated to safety. By isolating the safety objective, the effectiveness of our architecture is demonstrated through improved safety performance in experimental results.}
\item{A hybrid parameterized action space structure is designed to combine finer-grained guidance and control commands to adapt to hybrid road modality. This hybrid action space consists of discrete actions and their corresponding continuous parameters, which together generate both abstract guidance and concrete control outputs. Our design achieves greater driving flexibility and reduced behavioral fluctuations, ensuring compatibility between driving efficiency and action consistency.}
\item{An epistemic uncertainty-based exploration mechanism is developed to enhance learning efficiency and complement the hybrid action space structure. By dynamically adjusting the direction and magnitude of exploration according to uncertainty and its trends, the agent is encouraged to more rapidly explore high-uncertainty regions for potentially effective policies. This exploration mechanism significantly improves the learning efficiency of multi-objective compatible policies.}
\end{enumerate}

The remainder of this paper is organized as follows: Section~\ref{sec:Related Works} reviews related work, while Section ~\ref{sec:Methodology} outlines the methodology. Specific implementation details are presented in Section~\ref{sec:Implementation}. Section~\ref{sec:Results} discusses the experimental results, and the conclusions are provided in Section~\ref{sec:Conclusion}.

%% file: II_Related_Works.tex
\section{Related Works}\label{sec:Related Works}

The AD task involves making complex sequential decisions in a dynamic environment and can therefore be modeled as Markov Decision Processes (MDPs)~\cite{2024TNNLS_Wang, 2021TNNLS_Gao}. The MDP is commonly represented as a tuple $< {\cal S},{\cal A},{\cal R},{\cal T},\gamma  >$ , where ${\cal S}$ is the state space, ${\cal A}$ is the action space, ${\cal R}$ is the reward function, ${\cal T}$ is the transition function, and $\gamma$ is the discount factor. At time $t$, the RL agent selects action $a_t \in {\cal A}$ based on state $s_t \in {\cal S}$, then receives reward $r_t \in {\cal R}$ from the environment and transitions to state $s_{t+1}$ according to ${\cal T}$. The goal of the agent is to find an optimal policy through trial-and-error to maximize the expected reward.

\subsection{Multi-Objective Policy Evaluation}

For AD problems, multiple attitudes should be considered, requiring the policy to be multi-objective compatible. The objectives are sometimes conflicting, like safety and driving efficiency~\cite{OurSAE}. The most common design is to linearly combine all attributes into a single, additive reward function for policy evaluation~\cite{OurITSC}, typically based on mainstream RL algorithms such as Deep Q-Network (DQN)~\cite{DQN}and Soft Actor-Critic (SAC)~\cite{SAC}. Specifically, the weights of this linearly expressed reward function are typically determined through manual design after multiple trial-and-error iterations~\cite{2024ArXiv_Abouelazm}, or by applying Inverse-RL to human demonstrations~\cite{2023TITS_Wen}. However, policy evaluation under this linear assumption may be inaccurate because the highest rewarding action may not be the one that enables multi-objective compatible driving~\cite{2024NIPS_Cai, 2024NNLS_He}, leading to reduced policy performance~\cite{2016arXiv_Amodei}. Additionally, a single critic representing multiple attitude rewards forces the learning of value coherence, which may not accurately reflect the true critic and degrade policy quality~\cite{2022ICLR_Mysore}.

The MORL has recently attracted significant attention for its ability to address complex decision-making problems involving multiple, often conflicting, objectives~\cite{2024NNLS_He}. A typical MORL approach employs an architecture with multiple critics to enable multi-objective compatible policy updates~\cite{2024NIPS_Cai, 2021SAGE_Wang, 2023PartC_He, 2022arXiv_Srinivasan}. Specifically, key attributes are separated from a single reward function by defining multiple reward functions, with each attribute treated as an independent evaluation objective~\cite{2024NIPS_Cai}. Several AD-related studies have demonstrated the advantages of MORL in driving tasks by incorporating objectives such as safety~\cite{2021SAGE_Wang}, efficiency~\cite{2023TVT_Yang}, and comfort~\cite{2023PartC_He}. Additionally, \cite{2022arXiv_Srinivasan} introduces a pre-trained safe-critic to guide the policy towards safer actions. However, most prevalent MORL methods are derived from traditional RL algorithms and typically support only a single type of action output, which consequently limits the multi-objective compatibility of driving behaviors. Furthermore, such complex reward function designs require RL agents to consume significantly more time and computational resources during training to achieve sufficient exploration.Therefore, we propose HPA-MoEC, an advanced MORL architecture that integrates a novel hybrid action space and introduces epistemic uncertainty via ensemble-critic to enhance policy exploration and  learning capability. Compared to previous MORL methods, HPA-MoEC achieves better and faster multi-objective learning in hybrid road modality.

\subsection{Action Space Structure}

Many current RL-based AD methods use a single action type to control vehicle, which fail to be compatible with high driving flexibility and small behavior fluctuations. On one hand, some studies use a discrete action space to generate abstract behavior decisions, offering long-term targets that indirectly guide vehicle control. Specifically, \cite{2019SMC_Nageshrao, 2022TITS_Li} use DQN and its improved versions to generate semantic lateral actions, such as left or right lane changes. Additionally, \cite{2018IV_Wolf, 2023TVT_Chen} introduces longitudinal discrete acceleration and deceleration actions. To provide clearer guidance, some studies select from a discrete set of trajectories \cite{2020IV_Yu, OurAIAS} or directly generate the positions and desired speeds of target points \cite{2023ArXiv_Lu}. However, these methods often reduce the alignment between agent outputs and driving behavior, as they rely on integration with a basic controller for vehicle control, which limits flexibility. On the other hand, some studies \cite{2024ESA_Wang,2023TIV_Qi} directly generate steering angles laterally and accelerations longitudinally from a continuous action space, aiming to enhance flexibility. However, fluctuations in the network's output can cause frequent changes in steering angle and acceleration commands \cite{2025ICLR_Anonymous}. In scenarios with dedicated lanes, lateral fluctuations caused by steering angle variations will result in unpredictable paths. The experimental results in~\cite{2022TVT_Chen} provide further evidence for the existence of driving behavior fluctuations. In contrast, fluctuations in longitudinal acceleration are manageable and enable more flexible speed trajectories \cite{2021PartC_Guo}.

For compatibility of flexibility and small behavior fluctuations, several studies design hybrid actions by discretizing parts of a continuous action space \cite{2019IV_Wei, 2024TIV_Liu}, or using a parameterized action space \cite{2022TVT_Chen, 2021PartC_Guo, 2024Machines_Lin}, which generates both lateral discrete abstract targets and longitudinal continuous concrete acceleration commands. Additionally, \cite{2022TITS_Peng} designs a dual-layer decision-making control model that combines parallel DQN and Deep Deterministic Policy Gradient (DDPG) for hybrid output. \cite{2024CSL_Gurses} trains skill-agents for various driving objectives to output acceleration, from which DQN can flexibly select. However, the aforementioned studies fail to sufficiently integrate discrete and continuous actions, nor do they adequately account for the hybrid nature of road structures in complex driving environments. In contrast, our novel hybrid action space is specifically tailored to driving, providing abstract guidance compatible with hybrid road modality alongside continuous concrete control commands.


\subsection{Policy Exploration Mechanism}

Policy exploration helps agents discover potentially multi-objective compatible policies. A proper exploration mechanism can accelerate the learning process to converge to a viable policy faster~\cite{2023IS_Liu}. However, the most common exploration strategy in RL is random exploration, such as $\epsilon$-greedy in DQN~\cite{2010Acai_Tokic}, random noise in TD3~\cite{TD3}, and the maximum entropy mechanism in SAC~\cite{SAC}. This randomized mechanism makes policy exploration lack of orientation and leads to repeated collection of experience samples, which reduces the training efficiency~\cite{2022TNNLS_Chai}. Although some studies attempt to introduce reward novel state~\cite{2020AAAI_Machado} or error of reward~\cite{2021ICCAS_Usama} to modify the exploration level, this does not change the nature of random exploration. This inefficient exploration mechanism limits the potential performance of RL policies, particularly when pursuing multi-objective compatibility in complex traffic scenarios. Some other studies~\cite{2024TNNLS_Wu} attempt to use reward shaping to encourage exploration, but the manually imposed rewards heavily depend on the designer’s experience.

Some studies use model ensemble technique \cite{2020TSMCS_Gao, 2025TNNLS_Gao} to capture epistemic uncertainty and select actions that encourage the agent to explore high-uncertainty areas\cite{2020ICML_Zhang, 2024NIPS_Kim}, thus accelerating policy training. Few studies leverage epistemic uncertainty in AD tasks to improve driving policy training efficiency~\cite{2024TITS_Zhang}. Therefore, this paper develops an epistemic uncertainty-based exploration mechanism with multiple ensemble-critics for hybrid actions, enabling faster learning of multi-objective policy.

%% file: III_Methodology.tex
\begin{figure}[!tb]
  \centering
  \includegraphics[width=0.49\textwidth]{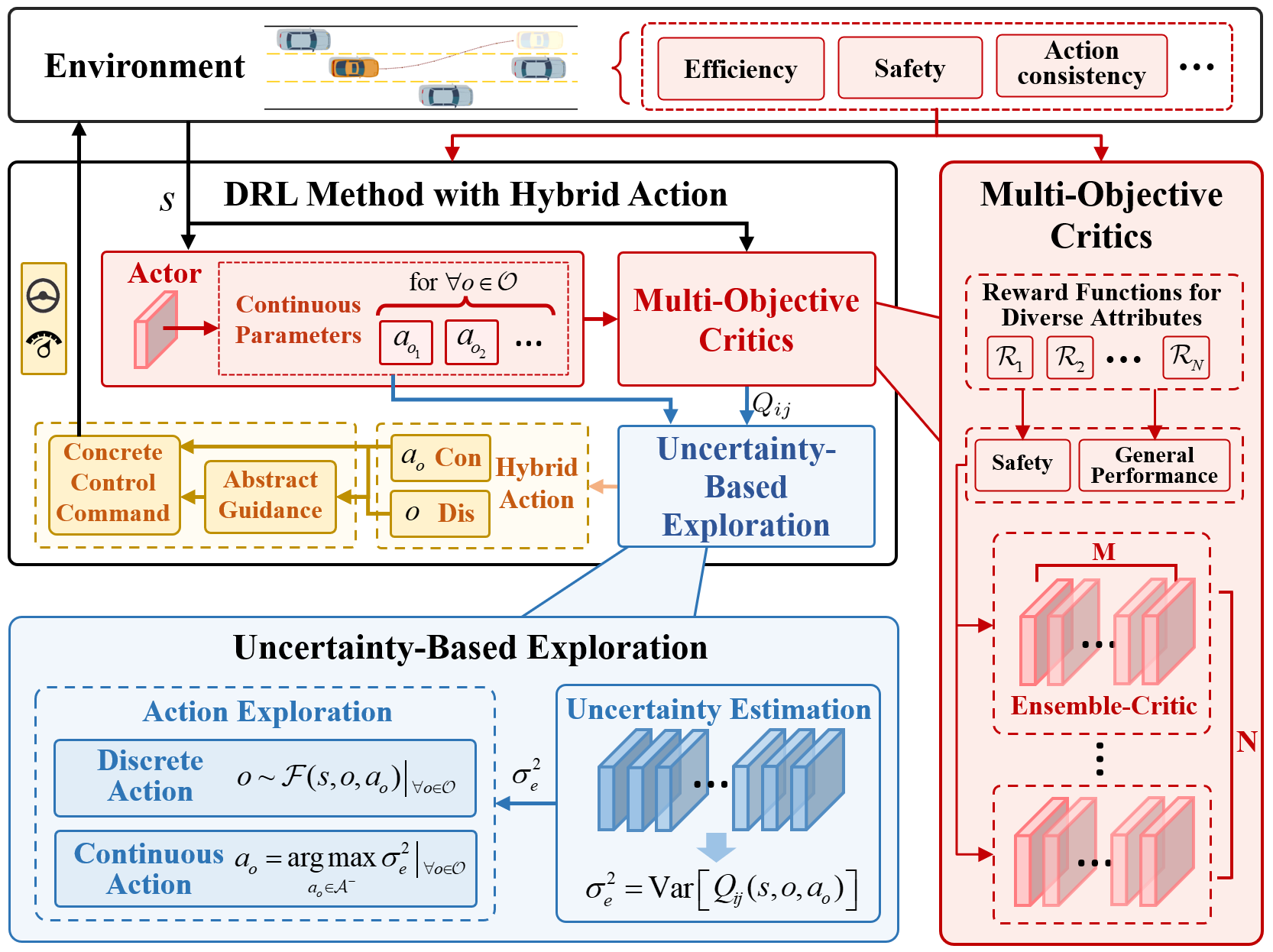}
  \caption{The overall framework of proposed HPA-MoEC. The actor of RL Method firstly generates the continuous action parameters $a_o$ based on states $s$, which are then input into the Multi-Objective Critics module along with $s$ for evaluating the value function. This module consists of $N$ ensemble-critics corresponding to the different attributes, and each of them is an ensemble of $M$ critics. The Exploration Strategy module then captures epistemic uncertainty from the ensemble-critics and selects the final hybrid action $(o, a_o)$ that enhances training efficiency.}
  \label{frame}
\end{figure}

\section{Methodology}\label{sec:Methodology}

In this section, we will present the overall framework and specific formulation details related to the HPA-MoEC methodology.

\subsection{Overall Framework}\label{sec:Methodology:A}

The method proposed in this paper is based on a hybrid parameterized action space for policy evaluation and improvement, considering multiple objectives to achieve multi-objective compatibility. Thus, the MDP can be rewritten as a new tuple $< {\cal S},{\cal H},\left[ {{{\cal R}_1}, \cdots ,{{\cal R}_N}} \right],{\cal T},\gamma  >$, where:

\begin{itemize}
\item{${\cal H}$ represents the hybrid parameterized action space, where ${\cal H} = \left\{ {\left( {o,{a_o}} \right)\left| {{a_o} \in {{\cal A}_{\cal O}},{\rm{for\text{  }}}\forall o \in {\cal O}} \right.} \right\}$. The $o$ is the discrete action option selected from the discrete action option set ${\cal O}$. The $a_o$ can be seen as the continuous action parameter corresponding to $o$, drawn from the continuous interval ${\cal A}_{\cal O}$ corresponding to ${\cal O}$.}
\item{$\left[ {{{\cal R}_1}, \cdots ,{{\cal R}_N}} \right]$ represents a set of $N$ reward functions, where ${\cal R}_i$ denotes the $i$-th reward function for $i \in~[1, \cdots ,N]$.}
\end{itemize}

To construct a fine-grained abstract guidance suitable for hybrid road modality, the designed hybrid action space enables the agent to simultaneously output discrete actions $o$ and continuous action parameters $a_o$, ensuring optimality in both. These outputs are then used to generate both abstract guidance and concrete control commands. Specifically, lateral concrete control commands are generated by combining abstract guidance with prior knowledge, while longitudinal commands are directly derived from $a_o$.

In addition, the agent should consider multiple attributes of the AD task during policy evaluation and efficiently explore multi-objective compatible viable policies. Therefore, we design the Multi-objective Ensemble-Critic framework, which takes $N$ attributes as evaluation objectives and helps agent explore in high-certainty regions. Specifically, the framework consists of $N$ ensemble-critics, which work together for policy evaluation based on the reward functions $\left[ {{{\cal R}_1}, \cdots ,{{\cal R}_N}} \right]$, each focusing on different attributes. Meanwhile, each ensemble-critic consists of $M$ critics. The epistemic uncertainty $\sigma_{e}$ and its change trend can be captured through ensemble-critic, which helps to orient exploration. The overall framework of the proposed HPA-MoEC method is shown in Fig.~\ref{frame} .

\subsection{Policy and Value Function Representation}

Under the hybrid parameterized action space, the state-action value function of the optimal policy can be described by the Bellman optimal equation as follows:
\begin{equation}
\label{Q_normal}
Q\!\left( {{s_t},{o_t},{a_{o,t}}} \right)\! = \!\mathbb{E} \!\left[ {{r_t}\! +\! \gamma \mathop {\max \!}\limits_{\!o \in {\cal O}} \left\{\! {\mathop {\sup \!}\limits_{{a_o} \in {{\cal A}_{\cal O}}} \!{Q^{'}} \! \left(\! {{s_{t + 1}},o,{a_o}}\! \right)\!}\! \right\}} \!\right]\!.
\end{equation}

HPA-MoEC consists of $N$ ensemble-critics, each composed of $M$ critics, resulting in a total of $N*M$ critics for value function evaluation. Specifically, each critic can estimate the value of the action $(o,a_o)$ in state $s$ based on its focused attributes. Let ${Q_{ij}}$ represents the optimal value function evaluated by the $i$-th critic corresponding to ${\cal R}_i$:
\begin{equation}
\label{Q_ij_1}
\begin{split}
{Q_{ij}}(\! {{s_t},{o_t},{a_{o,t}}}\!) \!=\! \mathbb{E} \!\left[\! {r_{i,t}\! +\! \gamma \mathop {\max }\limits_{o \in {\cal O}} \!\left\{\! {\mathop {\sup }\limits_{{a_o} \in {{\cal A}_{\cal O}}}\! {Q_{ij}}\!\left( {{s_{t + 1}},o,{a_o}} \right)\!}\! \right\}}\! \right]
\end{split}
\end{equation}
where $j \in [1, \cdots ,M]$, $r_{i,t} = {{\cal R}_i}(s,o,a_o)$. However, finding the optimal continuous action $a_o$ is challenging in a hybrid parameterized action space. To overcome this, we assume that the value function is fixed, meaning that for any state $s$ and discrete action $o$, the $a_o$ depends on state $s$. At this stage, the problem of optimizing in the continuous space becomes determining the mapping from state $s$ to action $a_o$: ${\cal S} \to {{\cal A}_{\cal O}}$. By using a deterministic policy network $\mu (s;{\theta ^\mu })$ to approximate this mapping, the continuous action $a_o$ can be obtained, with network parameters $\theta _\mu$. This policy network is known as the actor. Meanwhile, a value network is employed to approximate the value function $Q_{ij}$, with parameters $\theta _{ij}^Q$. Given the assumption that the value function is fixed, the MDP in the parameterized action space can be viewed as the process of exploring $\theta^{\mu}$ for a given $\theta _{ij}^Q$ :
\begin{equation}
\label{sup_to_mu}
{Q_{ij}}\left( {s,o,\mu \left( {s;{\theta ^\mu }} \right);\theta _{ij}^Q} \right) \approx \mathop {\sup }\limits_{{a_o} \in {{\cal A}_{\cal O}}} {Q_{ij}}\left( {s,o,{a_o};\theta _{ij}^Q} \right)\left| {_{\forall o \in {\cal O}}} \right..
\end{equation}
Specifically, this process can be approximated using a two-timescale update rule~\cite{1997SCL_Borkar}, where the training update step size for $\theta_{ij}^Q$ is much larger than that for $\theta_{ij}^\mu$. Therefore, $Q_{ij}$ can be expressed as:
\begin{equation}
\label{Q_ij_real}
\begin{split}
{Q_{ij}}&\left( {{s_t},{o_t},{a_{o,t}};{\theta _{ij}^Q}} \right) = \\ & \mathbb{E} \left[ {r_{i,t} + \gamma \mathop {\max }\limits_{o \in {\cal O}} {Q_{ij}}\left( {{s_{t + 1}},o,\mu \left( {{s_{t + 1}};{\theta ^\mu }} \right);\theta _{ij}^Q} \right)} \right].
\end{split}
\end{equation}
To pursue higher returns, referring to the value network update target in DQN~\cite{DQN}, the update target for a single critic is:
\begin{equation}
\label{y_ij}
\begin{split}
{y_{ij,t}} = r_{i,t} + \gamma \mathop {\max }\limits_{o \in {\cal O}} Q_{ij}'\left( {{s_{t + 1}},o,\mu '\left( {{s_{t + 1}};{\theta ^{\mu '}}} \right);\theta _{ij}^{Q'}} \right),
\end{split}
\end{equation}
where, ${Q_{ij}}'$ and $\mu'$ are the target networks used to assist in updating the critic and actor, with parameters $\theta _{ij}^{Q'}$ and $\theta _{ij}^{\mu '}$, respectively.

In our architecture, each critic is not updated independently. For each ensemble-critic, every critic within shares the same driving attribute. Then, all ensemble-critics collaborate to guide the actor in learning a multi-objective compatible driving policy. To ensure consistency among all critics in evaluating driving behavior, the evaluation results—both for specific attributes and overall performance—should be fed back to each critic, for updating networks. Therefore, it is necessary to construct the critic's update target from both the ensemble-critic perspective and the multi-objective compatible overall perspective. For the $i$-th ensemble-critic, its overall evaluation of the policy's performance under a given attribute is the expectation of the value provided by the $M$ critics:
\begin{equation}
\label{Q_i}
\begin{split}
{{\bar Q}_i}\left( {{s_t},o,{a_{o,t}}} \right) & = \mathbb{E} {_{j \in \left[ {1, \cdots ,M} \right]}}\! \left[ {{Q_{ij}}\!\left( {{s_t},o,\mu \left( {{s_{t + 1}};{\theta ^\mu }} \right);\theta _{ij}^Q} \right)} \right] \\
 & = \frac{1}{M}\!\sum\limits_{j = 1}^M \!{{Q_{ij}}\!\left( {{s_t},o,\mu \left( {{s_{t + 1}};{\theta ^\mu }} \right);\theta _{ij}^Q} \right)\!\left| {_{\forall o \in {\cal O}}} \right.}.
 \end{split}
\end{equation}
Correspondingly, the overall target of this ensemble-critic during training can be expressed as:
\begin{equation}
\label{y_i}
\begin{split}
{{\bar y}_{i,t}} = r_{i,t} + \gamma \mathop {\max }\limits_{o \in {\cal O}} {{\bar Q}_i}'\left( {{s_{t + 1}},o,{a_{o,t + 1}}} \right),
\end{split}
\end{equation}
where ${\bar Q_i}'$ is the expectation of all $Q_{ij}'$ for $j \in~[1, \cdots ,M]$, similar to~Eq.\eqref{Q_i}.

In addition, the actor's outputs assign different attention to the $N$ ensemble-critics, according to the weights $\omega_i$. Thus, the value function for evaluating the policy's multi-objective compatibility at the overall level can be represented as follows:
\begin{equation}
\label{Q_all}
\begin{split}
{Q_{all}}\left( {{s_t},o,{a_{o,t}}} \right) = \sum\limits_{i = 1}^N {{\omega _i}{{\bar Q}_i}\left( {{s_t},o,{a_{o,t}}} \right)} \left| {_{\forall o \in {\cal O}}} \right.,
\end{split}
\end{equation}
where  $\sum\nolimits_i^N {{\omega _i}} =1$. Building on this, the overall target for all critics in the HPA-MoEC can be written as:
\begin{equation}
\label{y_all}
\begin{split}
{y_{all,t}} = r_t^{all} + \gamma \mathop {\max }\limits_{o \in {\cal O}} Q_{all}'\left( {{s_{t + 1}},o,{a_{o,t + 1}}} \right),
\end{split}
\end{equation}
where $r^{all}$ combines the attribute rewards based on the attention level of each ensemble-critic, i.e., ${r^{all}} = \sum\nolimits_i^N {{\omega _i}{r_i}} $. The $Q_{all}'$ is denoted by the weighted sum of ${\bar Q_i}'$, similar to Eq.\eqref{Q_all}.

Thus, the update of the $\theta _{ij}^Q$ considers not only the critic's own TD error but also the average TD error of all critics in the ensemble for a given attribute, and the overall TD error of all critics. For these three aspects, corresponding loss functions are defined as follows:
\begin{equation}
\label{L_ij}
\begin{split}
{{\cal L}_{ij,t}}(\theta _{ij}^Q) = \frac{1}{2}{\left[ {{y_{ij,t}} - {Q_{ij}}\left( {{s_t},{o_t},{a_{o,t}};\theta _{ij}^Q} \right)} \right]^2},
\end{split}
\end{equation}
\begin{equation}
\label{L_i}
\begin{split}
{{\cal L}_{i,t}}(\theta _{ij}^Q) = \frac{1}{2}{\left[ {{y_{i,t}} - {{\bar Q}_i}\left( {{s_t},{o_t},{a_{o,t}}} \right)} \right]^2},
\end{split}
\end{equation}
\begin{equation}
\label{L_all}
\begin{split}
{{\cal L}_{all,t}}(\theta _{ij}^Q) = \frac{1}{2}{\left[ {{y_{all,t}} - {Q_{all}}\left( {{s_t},{o_t},{a_{o,t}}} \right)} \right]^2}.
\end{split}
\end{equation}
To prevent any critic from significantly deviating due to random factors and disrupting policy convergence, we have added a guiding term to the loss function of the parameter $\theta _{ij}^Q$. This helps ensure that all critics in the ensemble-critic are updated in a similar direction:
\begin{equation}
\label{L_conv}
\begin{split}
{{\cal L}_{conv,t}}(\theta _{ij}^Q) = \frac{1}{2}{\left[ {{Q_{ij}}\left( {{s_t},{o_t},{a_{o,t}};\theta _{ij}^Q} \right) - {{\bar Q}_i}\left( {{s_t},{o_t},{a_{o,t}}} \right)} \right]^2}.
\end{split}
\end{equation}
In summary, when updating the parameters $\theta _{ij}^Q$, the final loss function account for the four aspects discussed earlier:
\begin{equation}
\label{L_Q_ij}
\begin{split}
{{\cal L}_t}(\theta _{ij}^Q) = {{\bf{\lambda }}_t} \cdot {{\bf{L}}^\text{T} _t}.
\end{split}
\end{equation}
where ${{\bf{L}}_t} = \left[ {{{\cal L}_{ij,t}},{{\cal L}_{i,t}},{{\cal L}_{all,t}},{{\cal L}_{conv,t}}} \right]$ represents the vector of loss function and ${{\bf{\lambda}}_t} = \left[ {{\lambda _1},{\lambda _2},{\lambda _3},{\lambda _4}} \right]$ is the corresponding weight vector. By backpropagating the loss defined in Eq.~\ref{L_Q_ij}, the value network $Q_{ij}$ can be updated iteratively.

The target for updating the actor is more straightforward, i.e., finding a multi-objective compatible optimal policy by maximizing the overall value function:
\begin{equation}
\label{L_mu}
\begin{split}
{{\cal L}_t}\left( {{\theta ^\mu }} \right) =  - \frac{1}{M}\sum\limits_{i = 1}^N {{\omega _i}\sum\limits_{j = 1}^M {\sum\limits_{o \in {\cal O}} {{Q_{ij}}({s_t},o,\mu \left( {{s_{t + 1}};{\theta ^\mu }} \right);\theta _{ij}^Q)} } }.
\end{split}
\end{equation}

\begin{figure}[!tb]
  \centering
  \includegraphics[width=0.44\textwidth]{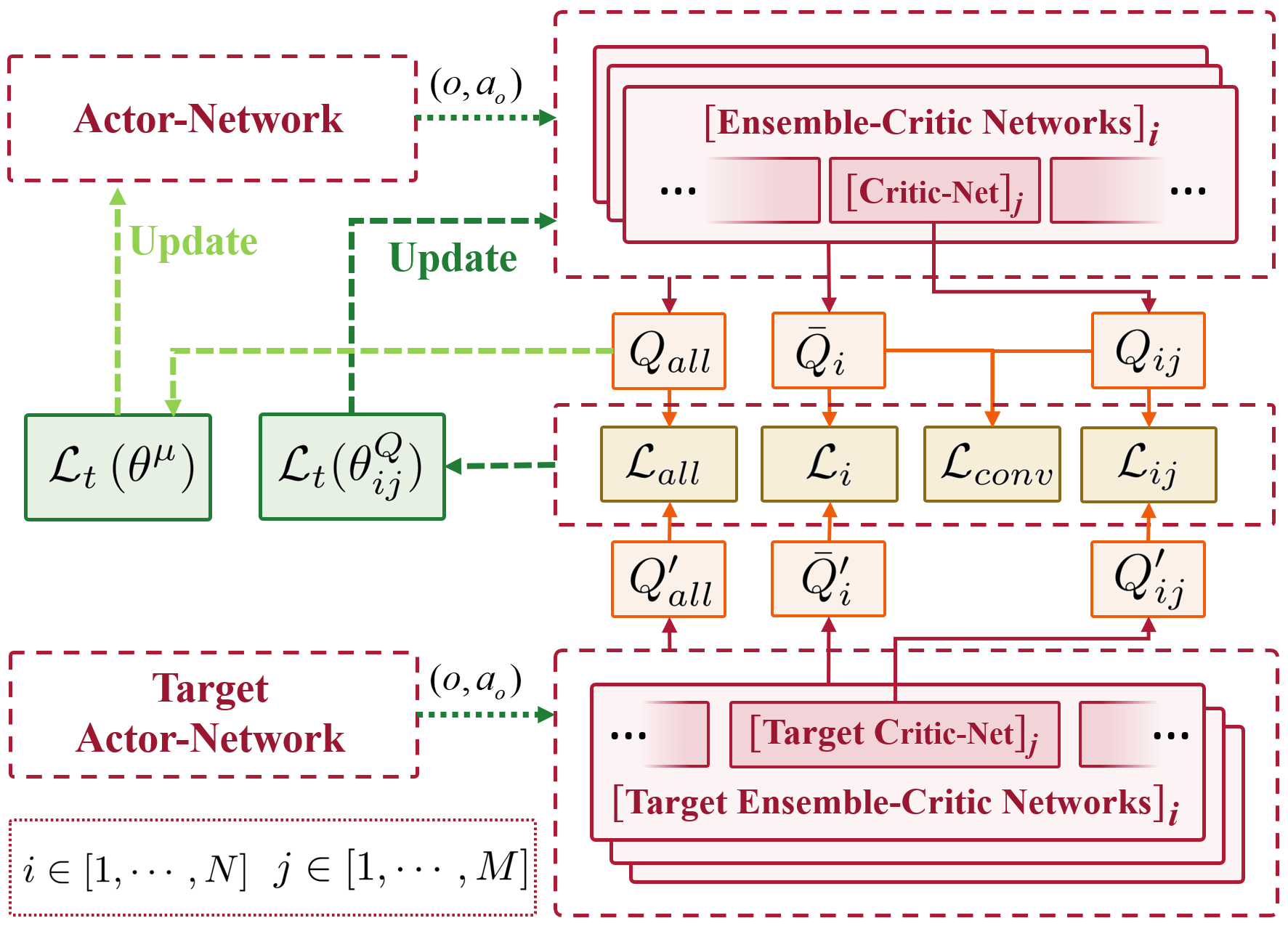}
  \caption{The network parameter update process for actor and any critic. The target networks are soft-updated.}
  \label{fig:networks update}
\end{figure}

Overall, the updating process of the actor's parameter $\theta ^\mu$ and any critic's parameter $\theta ^Q_{ij}$ is shown in Fig.~\ref{fig:networks update}.

\subsection{Uncertainty Estimation and Exploration Strategy}

Epistemic uncertainty reflects the agent's lack of knowledge due to incomplete learning and can be captured by ensemble-critic~\cite{2020IV_Hoel}. In the $i$-th ensemble-critic, a larger discrepancy between the evaluation results of the critics indicates higher epistemic uncertainty about the corresponding attribute. Such discrepancies can be quantified by the variance, so the epistemic uncertainty $\sigma _{e,i}^2$ of the $i$-th attribute is: 
\begin{equation}
\label{sigma_e,i}
\begin{split}
\sigma _{e,i}^2\left( {s,o,{a_o}} \right) = {\rm{Va}}{{\rm{r}}_{j \in [1, \cdots ,M]}}\left[ {{Q_{ij}}\left( {s,o,{a_o}} \right)} \right]\left| {_{\forall o \in {\cal O}}} \right..
\end{split}
\end{equation}

Considering that different attention levels are assigned to each ensemble-critic to achieve multi-objective compatibility, the weights $\omega_i$ are also used to compute the agent's total epistemic uncertainty:
\begin{equation}
\label{sigma_e}
\begin{split}
\sigma _e^2\left( {s,o,{a_o}} \right) = \sum\limits_{i = 1}^N {{\omega _i}\sigma _{e,i}^2\left( {s,o,{a_o}} \right)} \left| {_{\forall o \in {\cal O}}} \right..
\end{split}
\end{equation}
In the parameterized action space, $a_o$ is treated as a parameter of $o$. Thus, the change in epistemic uncertainty for any action pair $(o, a_o)$ can be captured by the gradient: 
\begin{equation}
\label{gradient}
\begin{split}
{\cal G} = {\nabla _{{a_o}}}\sigma _e^2\left( {s,o,{a_o}} \right)\left| {_{\forall {a_o} \sim \mu (s)}} \right..
\end{split}
\end{equation}
Additionally, it is necessary to clarify that $\sigma _{\rm{e}}^2(s,o,{a_o})$ represents the epistemic uncertainty of the state-action pair for $\forall o \in {\cal O}$, while the overall uncertainty of the environment at state $s$ is denoted as $\sigma _{\rm{e}}^2(s)$. Specifically, the two are related as follows:
\begin{equation}
\label{sigma_e_s}
\begin{split}
\sigma _e^2(s) = \mathbb{E} \left[ {\sigma _e^2(s,o,{a_o})} \right] \left| {_{\forall o \in {\cal O}}} \right. = \frac{1}{{\left| {\cal O} \right|}}\sum\limits_{o \in {\cal O}} {\sigma _e^2(s,o,{a_o})}.
\end{split}
\end{equation}

Oriented by the captured epistemic uncertainty, the agent employs two different exploration strategies for the discrete action $o$ and its corresponding continuous action $a_o$ while exploring potentially viable policies. For continuous action, the agent’s final executed $a_o$ is determined by both the actor's output and the chosen $o$. Thus, the ideal continuous action exploration strategy is to solve a nonlinear continuous optimization problem: $\arg {\max _{{a_o} \in {{\cal A}_{\cal O}}}}\sigma _e^2(s,o,{a_o})\left| {_{\forall o \in {\cal O}}} \right.$, to maximize exploration across all discrete actions $o$. However, solving this problem is computationally expensive and impractical for efficient policy training. Therefore, we choose a cheaper alternative by constructing a finite set of actions ${\cal A}^ -$, where ${\cal A}^ - \subset {\cal A}_{\cal O}$, based on the actor's origin output and epistemic uncertainty gradient. This discretizes the problem of selecting high-uncertainty actions in the continuous domain:
\begin{equation}
\label{A^-}
\begin{split}
{{\cal A}^ - } = \left\{ {{a_o}\left| {{a_o}{\rm{ = sa}}{{\rm{t}}_{{{\cal A}_{\cal O}}}}\left[ {\mu (s) + \frac{{k \cdot {\varsigma }}}{K} {\cal G}} \right],k \sim {\cal U}\left( {1,K} \right)} \right.} \right\},
\end{split}
\end{equation}
\[\]
\begin{equation}
\label{a_o_select}
\begin{split}
{a_o} = \arg {\max _{{a_o} \in {{\cal A}^ - }}}\sigma _e^2(s,o,{a_o})\left| {_{\forall o \in {\cal O}}} \right.,
\end{split}
\end{equation}
where ${\cal U}\left( {1,K} \right)$ denotes a uniform distribution over integers from 1 to $K$. The $\varsigma$ is a coefficient that decreases with training steps, where $\varsigma  \in \left( {0,1} \right)$, reflecting the agent's focus on exploring actions. This simplified approach enhances continuous action exploration with low computational cost.

Similarly, the most exploratory discrete action is the one that maximizes epistemic uncertainty: $\arg {\max _{o \in {\cal O}}}\sigma _e^2(s,o,{a_o})$. However, when the epistemic uncertainty of all discrete actions in the set $\cal O$ is low, relying on epistemic uncertainty to choose actions contributes little to strategy exploration, since the agent is already confident about all actions. Thus, we define an uncertainty threshold $\sigma _{e,th}^2$ to ensure the agent adopts a greedy strategy and maximizes reward when its uncertainty is low. Additionally, since the parameters of the critic-networks are randomly initialized and their outputs may fluctuate, the estimation of epistemic uncertainty has fluctuations. We use a probabilistic approach rather than directly selecting the action with maximum uncertainty. Specifically, similar to the \textit{Softmax} function, the probability of selecting a discrete action is based on its uncertainty value, with the total probability across all actions summing to 1. Therefore, the selection of discrete actions follows the function ${\cal F}$, where~$o \sim {\cal F}(s,o,{a_o})\left| {_{\forall o \in {\cal O}}} \right.$:
\begin{equation}
\label{o_select}
\begin{split}
{\cal F} = \left\{ {\begin{array}{*{20}{c}}
{o \sim \varepsilon \left( {s,o,{a_o}} \right)\left| {_{\forall o \in {\cal O}}} \right.}&{{\rm{if \text{ }}}{\varsigma}\sigma _e^2\left( s \right) > \sigma _{e,th}^2}\\
{\mathop {\arg \max }\limits_{o \in {\cal O}} {Q_{all}}\left( {s,o,{a_o}} \right)}&{else}
\end{array}} \right.,
\end{split}
\end{equation}
\begin{equation}
\label{epsilon}
\begin{split}
\varepsilon \left( {s,o,{a_o}} \right) = \frac{{{e^{\sigma _e^2(s,o,{a_o})}}}}{{\sum\nolimits_{o \in {\cal O}} {{e^{\sigma _e^2\left( {s,o,{a_o}} \right)}}} }}\left| {_{\forall o \in {\cal O}}} \right.,
\end{split}
\end{equation}
where $\varepsilon$ indicates the probability of choosing each action.

Based on the methods discussed above, we provide the complete algorithmic training process for our HPA-MoEC in Algorithm~\ref{algorithm}.

\begin{algorithm}
    \caption{Training process of proposed HPA-MoEC} \label{algorithm}
    \begin{algorithmic}[1] 
    \Require Step sizes $\{\alpha, \beta \}$ ,total training steps $T$, soft-update parameter~$\tau$, number of critics in ensemble-critic $M$, attribute weight $\omega$, loss function weight $\bf{\lambda}$.
        \State \textbf{Initialize}: networks $\{\{Q_{ij}\}, \mu, \{Q'_{ij}\}, \mu'\}$ with random parameters $\{\{\theta _{ij}^{Q}\}, \theta ^\mu, \{\theta _{ij}^{Q'}\}, \theta ^{\mu'}\}$ for $i \in~[1, \cdots ,N]$ and $j \in~[1, \cdots ,M]$, replay buffer size $D$, exploration parameter $\varsigma$.
        \For{$t = 0$ to $T$} 
            \State \text{Get state} $s_t$ {from environment.}
            \State Capture $\sigma _e^2$ and its gradient according to Eq.~\eqref{sigma_e,i}~\eqref{gradient}.
            \State Select $a_{o,t}$ for ${\forall o \in {\cal O}}$, according to Eq.~\eqref{a_o_select}.
            \State Select $o_t$ according to Eq.~\eqref{o_select}. 
            \State Generate abstract guidance by $o_t$ and $a_{{o,t}}$.
            \State Generate concrete control commands for EV.
            \State Get $s_{t + 1}$ and $r_{i,t}$ from environment, for $i\! \in\![1,\! \cdots \!,\!N]$.
            \State Store $\{s_t, (o_t, a_{o,t}), \left[r_{1,t}, \cdots, r_{N,t} \right], s_{t + 1})\}$ into $D$.
            \State Sample transitions randomly from $D$.
            \State Calculate ${{\cal L}_t}(\theta _{ij}^Q)$ for each critic, according to Eq.\eqref{L_Q_ij}.
            \State Update every critic network via gradient descent:
            \State  \hspace{3em} $\theta _{ij, t+1}^Q \leftarrow \theta _{ij, t}^Q - \alpha _t {\nabla }{{\cal L}_t}(\theta _{ij}^Q)$.
            \State Calculate ${{\cal L}_t}(\theta ^\mu)$ for actor, according to Eq.\eqref{L_mu}.
            \State Update actor network via gradient descent: 
            \State  \hspace{3em} $\theta _{ij, t+1}^\mu \leftarrow \theta _{ij, t}^\mu - \beta _t {\nabla }{{\cal L}_t}(\theta ^\mu)$.
            \State Soft-update every target critic network:
            \State  \hspace{3em} $\theta _{ij, t+1}^{Q'} \leftarrow {\tau}{\theta _{ij, t}^Q} + (1 - {\tau}){\theta _{ij, t}^{Q'}}$.
            \State Soft-update actor: 
            \State  \hspace{3em} $\theta _{t+1}^{\mu'} \leftarrow {\tau}{\theta _{t}^\mu} + (1 - {\tau}){\theta _{t}^{\mu'}}$.
            \State update $\varsigma$, $s_t \leftarrow s_{t + 1}$.
            \If{$s_t$ is terminal}
                \State Reset environment.
            \EndIf
        \EndFor
        \State \Return
    \end{algorithmic}
\end{algorithm}

%% file: IV_Implementation.tex
\section{Implementation}\label{sec:Implementation}

Multi-lane highway scenarios are both common and challenging, requiring driving policies that satisfy objectives such as efficiency, action consistency, and safety. This section presents the implementation details of HPA-MoEC in these scenarios, including the MDP formulation, training setup, and baseline models.

\subsection{MDP Formulation}

\subsubsection{State Space}

An appropriate state space representation is essential for effective policy learning. Specifically, the state space includes feature information about the Ego Vehicle (EV) and six surrounding vehicles (SVs) in its current and adjacent lanes:
\begin{equation}
\label{state}
\begin{split}
{\cal S} \buildrel \Delta \over = \left\{ {\begin{array}{*{20}{c}}
{{{\left[ {{ID}_{lane},x,y,\varphi ,{v_x},{v_y}} \right]}^\text{EV}},}\\
{\left[ {{p_n},\Delta {x_n},\Delta {y_n},{\varphi _n},\Delta {v_{x,n}},\Delta {v_{y,n}}} \right]_{n \in \left[ {1 \cdots 6} \right]}^\text{SVs}}
\end{array}} \right\},
\end{split}
\end{equation}
where the state of the EV in the road coordinate system consists of six variables: lane ID, longitudinal and lateral position, heading angle, and longitudinal and lateral velocity. For the $n$-th SV, the relevant information includes: a presence flag, longitudinal and lateral position relative to the EV, heading angle, and longitudinal and lateral velocity relative to the EV. Notably, the EV only monitors SVs within the longitudinal observation range $\Delta x \in \left[ { - 80 m,160 m} \right]$.

\subsubsection{Hybrid Parameterized Action Space}

For multi-lane scenarios with hybrid road modalities, we design explicit hybrid parameterized actions as follows: i) discrete semantic decision action $b$, ii) continuous parameter $l$ for constructing a guiding path, and iii) continuous acceleration command $acc$. The concrete correspondence is: $b \leftarrow o, (l, acc) \leftarrow a_o$. Specifically, $b$ is selected from a discrete set $\left\{ {LLC: - {w_r},RLC:{w_r},LK:0} \right\}$, where $w_r$ represents the road width, with $LLC$ and $RLC$ representing left and right lane-change, respectively, and $LK$ indicating lane-keeping. Considering the vehicle kinematic model~\cite{OurITSC}, the value range for $l$ is defined as follows:
\begin{equation}
\label{con_action}
\begin{split}
l \in \left[ {\min \left( {\sqrt {4{R_0}{w_r} - w_r^2} ,\frac{{v_x^2}}{{2acc_{\max }^ - }}} \right),{e^{\left| {{v_x}} \right| + {w_r}}}} \right],
\end{split}
\end{equation}
where $R_0$ and $acc_\text{max}^ -$ represent the minimum turning radius and maximum braking acceleration of the EV. In addition, the range of the acceleration command $acc$ is $\left[ { - 3{\rm{m/}}{{\rm{s}}^2},3{\rm{m/}}{{\rm{s}}^2}} \right]$.

At each time step $t$, with $(b_t, l_t, acc_t)$ output by the agent, the positions of the guiding path points can be generated using a polynomial curve-based formula:
\begin{equation}
\label{polynomial curve}
\begin{split}
{y_{0,t + k}} = \sum\limits_{m = 0}^5 {{\gamma _m}x_{0,t + h}^m} ,{\text{  where  }}h \in \left[ {1, \cdots ,{H_p}} \right],
\end{split}
\end{equation}
where $(x_{t+h}, y_{t+h})$ represents the position of the point at time step $t+h$, and $H_p$ is the planning horizon. The coefficients $\gamma_m$ of the polynomial curve can be obtained by solving a system of linear equations. Specifically, the EV's position and heading at the starting point are known, while the heading at the end point can be obtained from the road information~\cite{OurSAE}. Actually, the guiding path is determined by the selection of its endpoint position $(x_{t+H_p}, y_{t+H_p})$, which is derived from the RL agent's output, where $x_{t+H_p} = l_t$ and $y_{t+H_p} = b_t$. As the guiding path generated, the EV's steering angle command $\delta$ is output using prior knowledge, specifically the Stanley algorithm in this paper. Finally, both the steering angle $\delta$ and the acceleration $acc$ are used together for EV driving control.

\subsubsection{Reward Function for Multi-Objective}

Since safety is the fundamental requirement for driving, safety attribute is treated as a distinct objective and a corresponding safety reward function is designed for one ensemble-critic. Other attributes are combined into a single general performance reward function for another ensemble-critic. This enhances the RL agent's compatibility with safety and general driving performance.

The safety reward function, ${\cal R}_{safe}$, focuses on safety in two aspects:
\begin{equation}
\label{R_safe}
\begin{split}
{{\cal R}_{safe}} =  - 10{f_{unsafe}} + 0.5{\rm{sa}}{{\rm{t}}_{\left[ {0,1} \right]}}\left[ {\frac{{\Delta t}}{{{t_{\max }}}}} \right],
\end{split}
\end{equation}
where, $f_{unsafe}$ is set to 1 when the EV goes off the road or collides with SVs, and 0 otherwise. To further identify potential safety risks, the safety reward function also includes the TTC (Time to Collision) metric, where $\Delta t$ is the estimated time to collision between the EV and the vehicle ahead, and $t_{max}$ is the maximum time for TTC evaluation. The values 10 and 0.5 are the weights assigned to the two aspects mentioned above, respectively.

The general performance reward function, ${\cal R}_{gen}$, incorporates considerations of efficiency, comfort, and interaction:
\begin{equation}
\label{R_gen}
\begin{split}
&{{\cal R}_{gen}} = {{\cal R}_{eff}} + {{\cal R}_{comf}} + {{\cal R}_{int}}\\
&{{\cal R}_{eff}} = \frac{{\left| {v - {v_t}} \right|}}{{{v_t}}} - \max (0,\frac{{{v_p} - v}}{{{v_p}}})\\
&{{\cal R}_{comf}} =  - 0.5\frac{{\left| \delta  \right|}}{{\left| {{\delta _{\max }}} \right|}} - 0.5\frac{{\left| {acc} \right|}}{{\left| {{acc_{\max }}} \right|}}\\
&{{\cal R}_{int}} =  - 0.1\sum\limits_{n = 1}^6 {\frac{{\left| {acc_n^{SV}} \right|}}{{\left| {{acc_{\max }}} \right|}}} ,
\end{split}
\end{equation}
where ${\cal R}_{eff}$ is efficiency reward, encouraging the EV to maintain a speed close to the target $v_t$. Meanwhile, a low-speed penalty is applied to minimize the impact of the vehicle's deceleration on overall traffic flow, with the threshold set at $v_p$. The ${\cal R}_{comf}$ is comfort reward, related to the action consistency of steering angle and acceleration, where $\delta_{max}$ and $acc_{max}$ denote the maximum values of the two control commands. Moreover, ${\cal R}_{int}$ represents the interaction reward, penalizing EV's interference with SV's motion while interacting with environment. The $acc_n^{SV}$ denotes the observed acceleration of the $n$-th SV. The number before each item is the weight of the attention given to it.

\subsection{Training Setup}

\begin{figure}[!tb]
  \centering
  \includegraphics[width=0.45\textwidth]{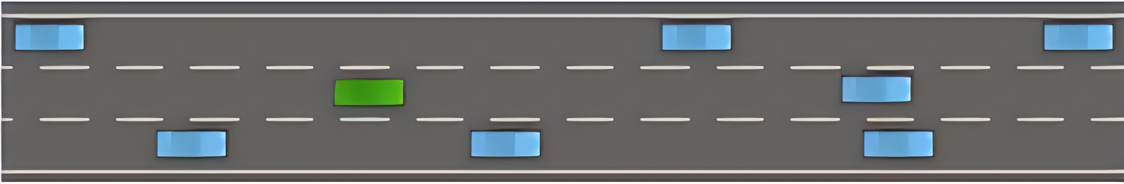}
  \caption{Schematic diagram of multi-lane highway environments in highway-env. Green vehicles represent EVs, while blue vehicles represent SVs.}
  \label{fig:highway-env}
\end{figure}

We developed a three-lane structured road environment using the AD simulation platform, highway-env~\cite{2018highway_env}, in which the EV attempts to accomplish a multi-objective compatible driving task. A schematic diagram of the study scenario is shown in Fig.~\ref{fig:highway-env}. Specifically, all vehicles, including the EV, are randomly placed on the three-lane road with random initial speeds. The IDM and MOBIL models are applied to control the longitudinal and lateral movements of the SVs~\cite{OurITSC}. The SVs may change lanes at appropriate times to get closer to the target speed, potentially disrupting the EV. Additionally, we use the vehicle capacity (V/C) to represent traffic congestion, setting it to 0.5 to create moderate congestion. This ensures the EV has enough space to change lanes without oversimplifying the environment.

During training, the episode ends when $f_{unsafe} = 1$, after which the environment and all vehicles are reinitialized. Each episode is capped at 200 seconds to avoid the EV operating for long periods in low-variability scenarios. Details of the hyperparameter settings used in the algorithm training are provided in Table~\ref{Hyperparameters}.

\begin{table}[ht]
\caption{Hyperparameters}
\vspace{-2mm}
\label{Hyperparameters}
\begin{center}
\begin{tabular}{|c||c||c|}
\hline
\textbf{Para.} & \textbf{Item} & \textbf{Value}\\
\hline
$M$ & Number of critics in an ensemble-critic & 6\\
\hline
$\omega_1 , \omega_2$ & Weights of ${\cal R}_{safe}$ and ${\cal R}_{gen}$ & 0.4, 0.6\\
\hline
$\gamma$ & Discount factor & 0.9\\
\hline
$\alpha$ & Training step size of critic & 0.01\\
\hline
$\beta$ & Training step size of actor & 0.001\\
\hline
$\lambda$ & Weights of loss functions for critic & [0.5,0.2,0.2,0.1]\\
\hline
$K$ & parameters for con-action exploration & 10\\
\hline
$\tau$ & Soft-update parameter & 0.005\\
\hline
$T$ & Number of steps for training &  200000\\
\hline
$\varsigma$ & Exploration weight parameter &  1$\to$0.001\\
\hline
$-$ & Number of hidden layers in critic/actor & 3\\
\hline
$-$ & Hidden layer size & 256\\
\hline
$-$ & Activation function & Tanh\\
\hline
$-$ & Replay buffer size & 40000 \\
\hline
$-$ & Sample batch size & 256\\
\hline
$-$ & Training optimizer & Adam\\
\hline
\end{tabular}
\end{center}
\vspace{-2mm}
\end{table}

Additionally, our method is tested on 200 episodes in both the training environment and the HighD~\cite{2018HighD} real-world dataset. For testing on the HighD dataset, the trained agent controls randomly selected vehicles, while the SVs follow their predefined trajectories.

\subsection{Comparison Models}

\subsubsection{Comparison Baseline}

To comprehensively evaluate the proposed HPA-MoEC, we compare it with several widely used RL methods for the AD task. All methods share the same training and testing environments, as well as the state space. The main difference is that, unlike HPA-MoEC, the other methods couple the attributes into a single reward function: ${{\cal R}_{base}} = {\omega _1}{{\cal R}_{safe}} + {\omega _2}{{\cal R}_{gen}}$. More importantly, the action spaces structure and policy exploration strategies in the following methods differ:
\begin{itemize}
\item{Deep Q-Network (\textbf{DQN})~\cite{DQN}: It only generates discrete semantic decisions and is paired with a PID controller to control the EV. The exploration strategy used is $\epsilon$-greedy.}
\item{SAC with Continuous actions (\textbf{SAC-C})~\cite{SAC}: It only outputs continuous control commands, which are lateral steering angle and longitudinal acceleration. Its exploration is enhanced through maximum entropy and the addition of Gaussian noise to the actions.}
\item{SAC with Hybrid actions (\textbf{SAC-H})~\cite{SAC}: SAC-H discretizes part of the continuous action space in SAC, producing outputs similar to HPA-MoEC.}
\item{PPO with Hybrid actions (\textbf{PPO-H})~\cite{PPO}: An on-policy actor-critic algorithm with action space similar to \mbox{SAC-H}.}
\end{itemize}
To ensure fair comparisons and reliable conclusions, all methods use the same network architecture, learning rate, and other key hyperparameters. Additionally, for method-specific parameters, we perform extensive tuning within reasonable ranges and select the optimal configuration for each method.

\subsubsection{Ablation Model}

To further validate the effectiveness of the three key techniques used in HPA-MoEC: i) Hybrid parametric action space structure; ii) Multi-critic policy evaluation architecture; and iii) Epistemic uncertainty-based policy exploration, we design the following ablation baselines:
\begin{itemize}
\item{\textbf{HPA-MoEC}: The method proposed in this paper includes all three technical components.}
\item{\textbf{HPA-Mo}: By removing component iii from HPA-MoEC, policy exploration is no longer oriented by uncertainty. Policy evaluation for each objective is performed by a single critic only, rather than by an ensemble-critic.}
\item{\textbf{HPA}: By removing component ii from HPA-Mo, only one overall objective remains, with one corresponding critic for evaluating policies that considers multiple attributes. In fact, this baseline is similar to the algorithm in~\cite{OurITSC}.}
\item{\textbf{DA-Mo}: By removing component i from HPA-Mo, this baseline generates only coarse-grained discrete semantic decisions as abstract guidance, which are combined with the PID controller to output steering angles. In fact, this baseline is similar to part of the work in~\cite{2024NNLS_He}.}
\end{itemize}

\subsection{Evaluation Metrics}

To evaluate the driving performance of the proposed method across multiple objectives, we used several metrics for each episode:
\begin{itemize}
\item{Average Reward (\textbf{AR}): AR is the ratio of total reward to episode length, offering a comprehensive evaluation of the RL agent's performance.}
\item{Collision Rate (\textbf{CR, \%}): Collisions result from hazardous driving behavior and can be used to evaluate the safety of the agent's driving policy.}
\item{Average Speed (\textbf{AS, m/s}): The EV's speed indicates the agent's ability to intelligently execute lane changes actions to enhance driving efficiency.}
\item{Number of Lane-change (\textbf{NL}): NL partially reflects the EV's flexibility and can be analyzed alongside AS to explain the reasons for improved driving efficiency.}
\item{Variance of Steering angle (\textbf{VS, $\text{rad}^2$}) and Acceleration (\textbf{VA, ${\text{m}^2}/{\text{s}^4}$}): VS and VA respectively indicate the vehicle's fluctuations in lateral and longitudinal behavior, reflecting the consistency of the driving policy's actions.}
\end{itemize}

%% file: V_Results_and_Discussions.tex
\section{Results and Discussions}\label{sec:Results}

\subsection{Training Performance}

\begin{figure}[!tb]
  \centering
  \includegraphics[width=0.47\textwidth]{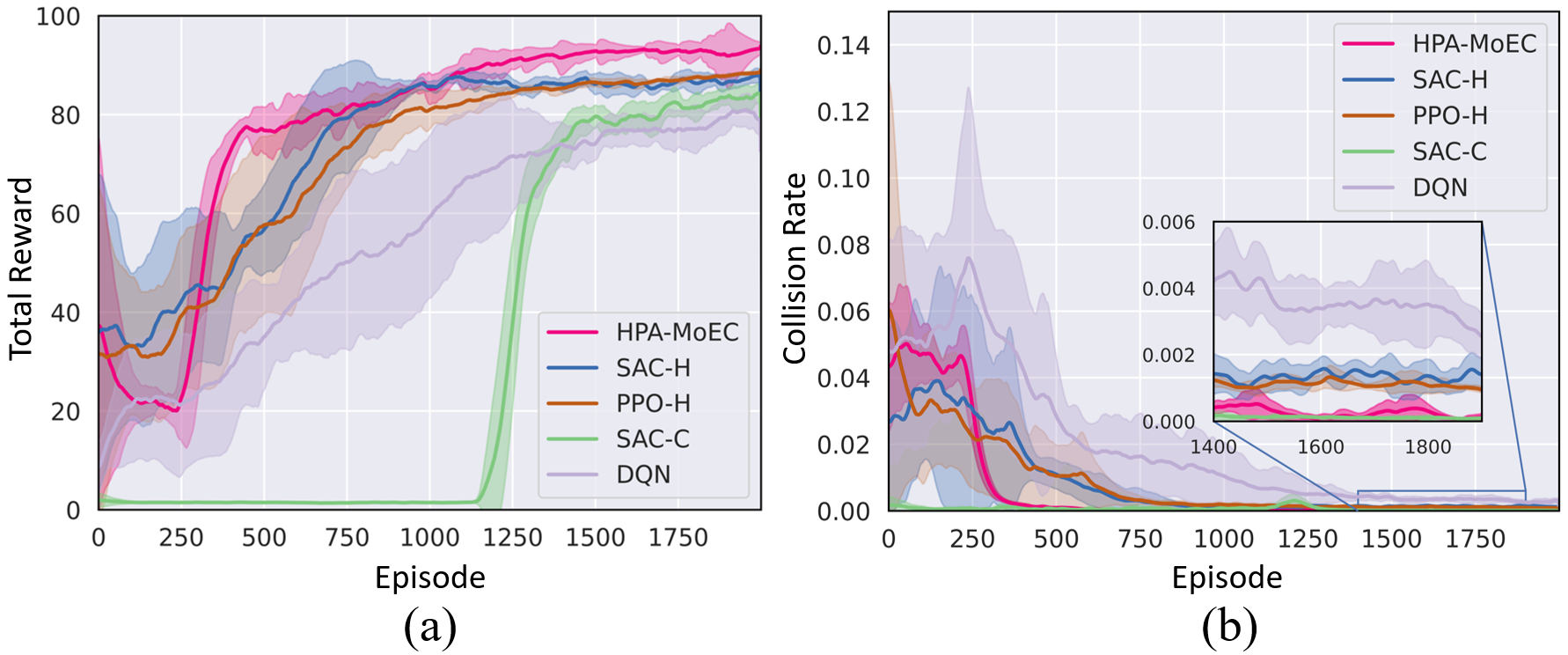}
  \caption{The training process of our method with comparison methods quantified by: a) Total Reward and b) Collision Rate.}
  \label{fig:train comp}
\end{figure}

The learning curves for general performance and safety during training are shown in Fig.~\ref{fig:train comp}, with each algorithm trained six times using different seeds. The total reward curve and corresponding variance distribution in Fig.~\ref{fig:train comp}(a) show that the our HPA-MoEC achieves higher rewards with smaller policy fluctuations. This indicates that, regardless of seed variations, its policy consistently converges to the best general performance. By comparison, the similar rewards achieved by SAC-H and PPO-H indicate that both of them perform worse than HPA-MoEC. Furthermore, without finer-grained guiding paths, the reward during SAC-C convergence is much lower, indicating poorer driving performance when both longitudinal and lateral direct control commands are output together. Using only semantic decision actions, the DQN receives the lowest reward, indicating that discrete actions alone are insufficient for complex driving tasks.

Additionally, once the minimum sample size required for training is gathered in the experience replay pool, the policy improvement speed of HPA-MoEC is significantly faster than that of all the baselines. This increase in training efficiency is attributed to the introduction of an epistemic uncertainty-based exploration strategy, which enables a oriented and faster exploration of potentially viable policies. Notably, since SAC-C directly controls the EV by outputting steering angle commands, it often veers off the road and ends the episode early, causing the reward curve to differ significantly from other methods.

As shown in Fig.~\ref{fig:train comp}(b), the change in CR for each method during training is illustrated, with the zoomed-in view of the converged curves highlighting that HPA-MoEC ultimately maintains a low CR. Thanks to the decoupling of the safety objective from the general performance objective within the multi-objective policy evaluation architecture, the agent places greater emphasis on safety. In contrast, \mbox{SAC-H} and PPO-H have slightly higher CRs, whereas DQN has the highest. Notably, although SAC-C performs poorly in total reward, it prioritizes the safety of the EV by maintaining a very low CR. This results from its conservative following behavior, which will be discussed in detail in Section~\ref{sec:Results:Testing:r}.

\subsection{Testing Performance}

\subsubsection{Testing with Rule-Based SVs} \label{sec:Results:Testing:r}

\begin{figure}[!tb]
  \centering
  \includegraphics[width=0.47\textwidth]{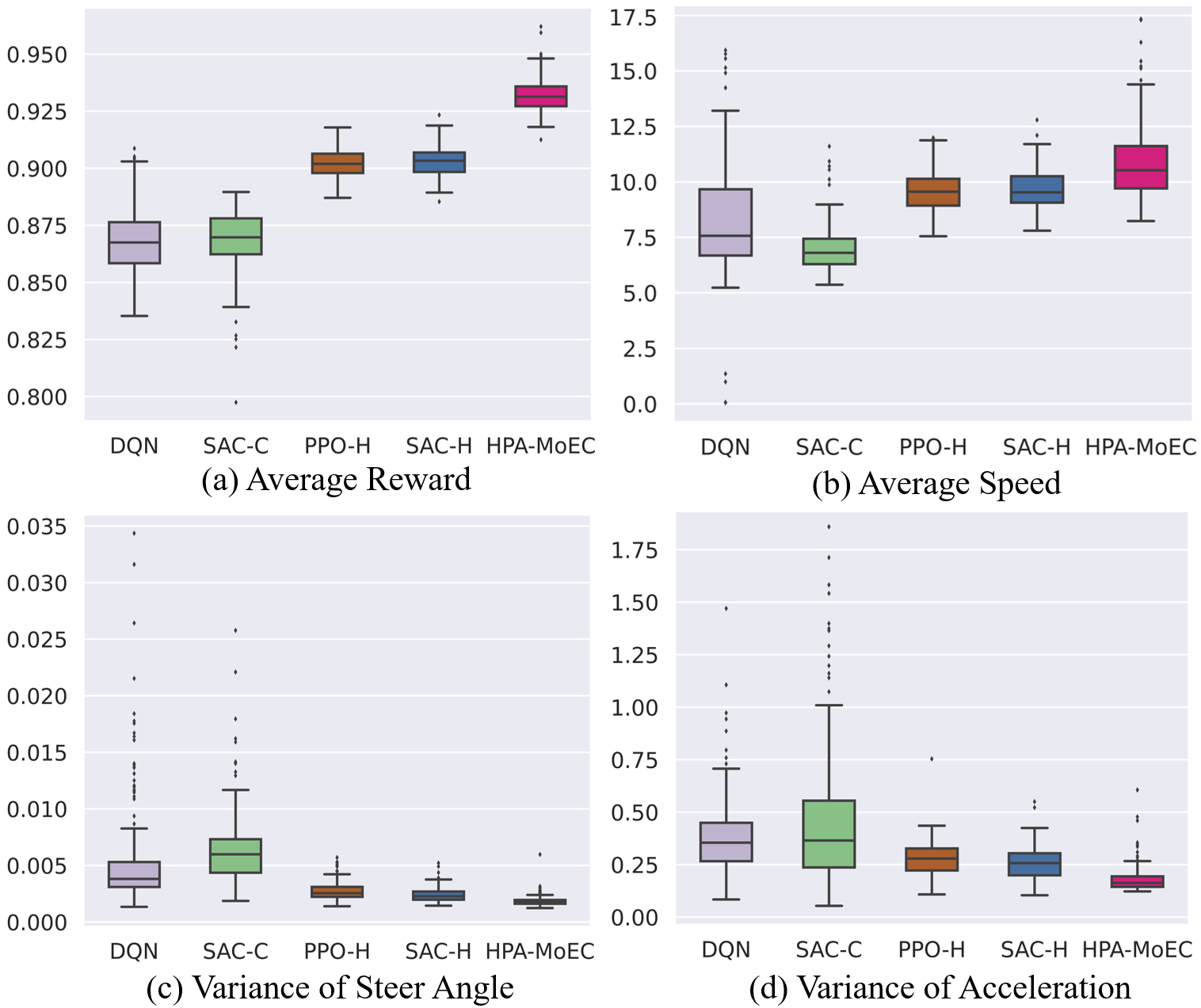}
  \caption{Metrics distribution of testing with Rule-Based SVs: (a) average reward, (b) average speed, (c) variance of steering angle, (d) variance of acceleration.}
  \label{fig:test comp}
\end{figure}

\begin{table}[!tb]  
\caption{Test results with rule-based SVs}
\centering  
\begin{threeparttable}
\begin{tabular}{c|cccccc}  
\toprule  
{\textbf{Method}} & \textbf{AR} & \textbf{AS} & \textbf{NL} & \textbf{VS} & \textbf{VA} & \textbf{CR} \\
\midrule  
DQN      & 0.860      & 8.18        & 7.71        & 0.0055      & 0.381       & 0.38\%       \\
SAC-C    & 0.868      & 6.95        & 2.04        & 0.0063      & 0.452       & 0.01\%       \\
PPO-H    & 0.902      & 9.57        & 5.76        & 0.0027      & 0.279       & 0.13\%       \\
SAC-H    & 0.903      & 9.62        & 5.57        & 0.0024      & 0.256       & 0.12\%       \\     
HPA-MoEC & 0.932      & 10.87       & 7.14        & 0.0019      & 0.181       & 0.04\%       \\       
\bottomrule  
\end{tabular}
\begin{tablenotes}
\footnotesize
\item[*] All values are averaged over 200 episodes.
\end{tablenotes}
\end{threeparttable}
\label{table:test comp}
\end{table}

The boxplots in Fig.~\ref{fig:test comp} illustrate the distribution of four metrics in testing: average reward (Fig.~\ref{fig:test comp}(a)), average speed (Fig.~\ref{fig:test comp}(b)), and the variance of steering angle and acceleration (Fig.~\ref{fig:test comp}(c) and Fig.~\ref{fig:test comp}(d)). The quantitative statistics for all metrics are provided in Table~\ref{table:test comp}. It can be observed that AS and CR jointly influence AR: a larger AR is achieved only by accommodating higher AS and lower CR, indicating that a superior driving policy must be compatible with both safety and traffic efficiency. Although there is no direct correlation between NL and AS, together they comprehensively reflect the driving style related to traffic efficiency. Furthermore, algorithms exhibiting higher VS typically also possess higher VA, suggesting that in most cases, the fluctuations in control commands tend to be consistent across both lateral and longitudinal dimensions. Specifically, the driving policy of the proposed HPA-MoEC demonstrates advantages in driving efficiency, action consistency, and safety.

In general, HPA-MoEC receives the highest AR, which is consistent with the training results and indicates a more effective driving policy. SAC-H and PPO-H also perform well, with similar AR levels. In contrast, the driving policy of DQN and SAC-C perform poorly and exhibit considerable fluctuation, with lower and more dispersed AR values.

\textbf{For driving efficiency}, HPA-MoEC achieves the highest AS through more flexible lane changes. In comparison, SAC-H and PPO-H have lower ASs due to reduced lane-changing flexibility, leading to suboptimal efficiency. Specifically, compared to SAC-H, HPA-MoEC improves AS by 13\% and increases NL by 28\%. Compared to the above methods with hybrid actions, SAC-C's direct control of the EV results in the lowest AS and the fewest NL, indicating its inability to effectively leverage lane-changing opportunities to increase speed. Relying on discrete actions, DQN achieves higher AS in some episodes by frequent lane changes, but its overall AS ranks second to last. Overall, the hybrid actions provide greater flexibility and thus improve driving efficiency, especially by using a parameterized action space to generate outputs rather than discretizing part of the continuous actions.

\begin{figure}[!tb]
  \centering
  \includegraphics[width=0.47\textwidth]{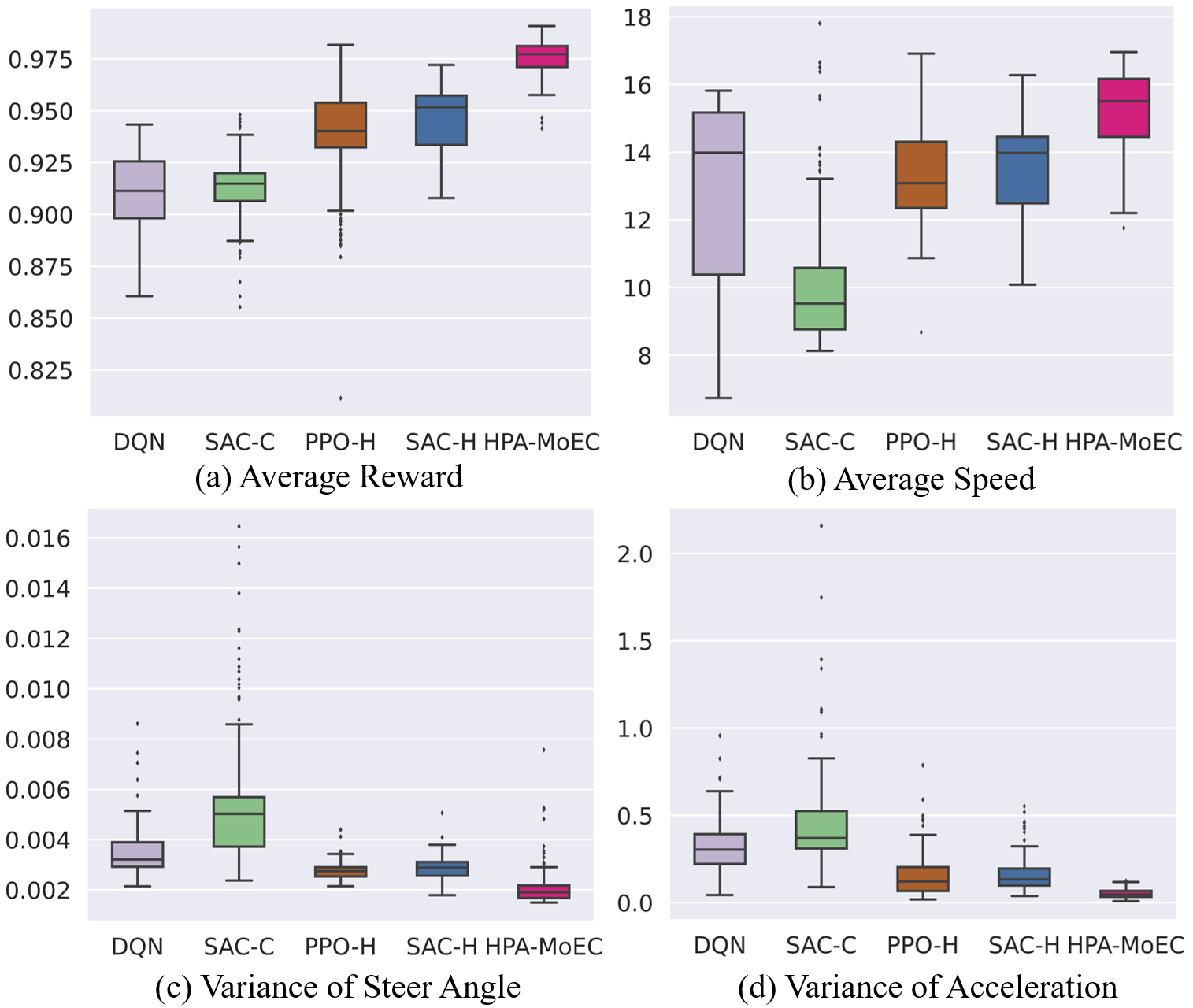}
  \caption{Metrics distribution of testing in HighD dataset.}
  \label{fig:highd comp}
\end{figure}

\begin{table}[!tb]  
\caption{Test results in HighD dataset}
\centering  
\begin{threeparttable}
\begin{tabular}{c|cccccc}  
\toprule  
{\textbf{Method}} & \textbf{AR} & \textbf{AS} & \textbf{NL} & \textbf{VS} & \textbf{VA} & \textbf{CR} \\
\midrule  
DQN         & 0.909      & 12.74       & 8.84        & 0.0035      & 0.316       & 0.29\%       \\
SAC-C       & 0.913      & 10.12       & 1.25        & 0.0054      & 0.451       & 0.00\%       \\
PPO-H       & 0.938      & 13.32       & 5.67        & 0.0028      & 0.155       & 0.04\%       \\
SAC-H       & 0.945      & 13.58       & 5.69        & 0.0029      & 0.160       & 0.05\%       \\
HPA-MoEC    & 0.976      & 15.27       & 7.03        & 0.0021      & 0.051       & 0.01\%       \\       
\bottomrule  
\end{tabular}
\begin{tablenotes}
\footnotesize
\item[*] All values are averaged over 200 episodes.
\end{tablenotes}
\end{threeparttable}
\label{table:highd comp}
\end{table}

\textbf{For action consistency}, HPA-MoEC exhibits the smallest VS and VA, implying a significant reduction in lateral and longitudinal driving behavior fluctuations. In comparison, although PPO-H and SAC-H also generate hybrid actions, their VS increases by 26\% and 42\%, respectively, while their VA increases by 41\% and 54\%, respectively. This indicates that the HPA-MoEC generates smoother guiding paths and acceleration commands through its parameterized action space. Notably, both SAC-C and DQN exhibit large VS and VA, indicating large behavior fluctuations. For DQN, the discrete decision set hampers smooth steering adjustments during lane changes and restricts acceleration flexibility. For SAC-C, the coupling between steering angle and acceleration commands makes it extremely challenging to produce smooth and regular outputs when both exhibit fluctuations.

\textbf{For safety performance}, HPA-MoEC demonstrates the second-lowest CR, trailing only to SAC-C, highlighting its strong focus on safety. This is facilitated by a policy evaluation design with safety attribute as a separate objective, achieving a CR reduction of 67\% and 69\% for HPA compared to SAC-H and PPO-H, respectively. Notably, since SAC-C directly outputs control commands through a single continuous action space, it learns an over-conservative driving policy. Although this extreme concern for short-term safety significantly reduces CR, it greatly sacrifices efficiency and action consistency. In contrast, the slight compromise in safety offered by HPA-MoEC brings significant improvements in efficiency and action consistency, which is more aligned with the multi-objective requirements of AD. Additionally, DQN has a CR of 0.38\%, much higher than other methods. With an average of 7.73 NL per episode, this indicates that its more aggressive driving policy increases the risk of putting the EV in danger.

\subsubsection{Testing in HighD-Dataset}

The testing results on the HighD dataset, including the distribution of evaluation metrics and quantitative statistics, are shown in Fig.~\ref{fig:highd comp} and Table~\ref{table:highd comp}, respectively. Compared to the constructed simulation scenario, the traffic density in HighD is sparser, and all methods demonstrate better driving performance. Clearly, HPA-MoEC still achieves the highest AR, showing the good adaptability of its driving policy. It also maintains excellent control over acceleration and flexible lane-changing abilities, resulting in the highest AS and the most NL, except for DQN. Additionally, the guiding path still plays a role in the reduction of vehicle behavior fluctuations, keeping the VS and VA low. In terms of safety, the emphasis on safety attributes in HPA-MoEC reduces the CR to just 0.01\%. Overall, HPA-MoEC outperforms all other baselines in terms of compatibility with the objectives of driving efficiency, action consistency, and safety, offering greater potential for real-world traffic applications.

\subsection{Ablation study}

\subsubsection{Training Performance}

\begin{figure}[!tb]
  \centering
  \includegraphics[width=0.47\textwidth]{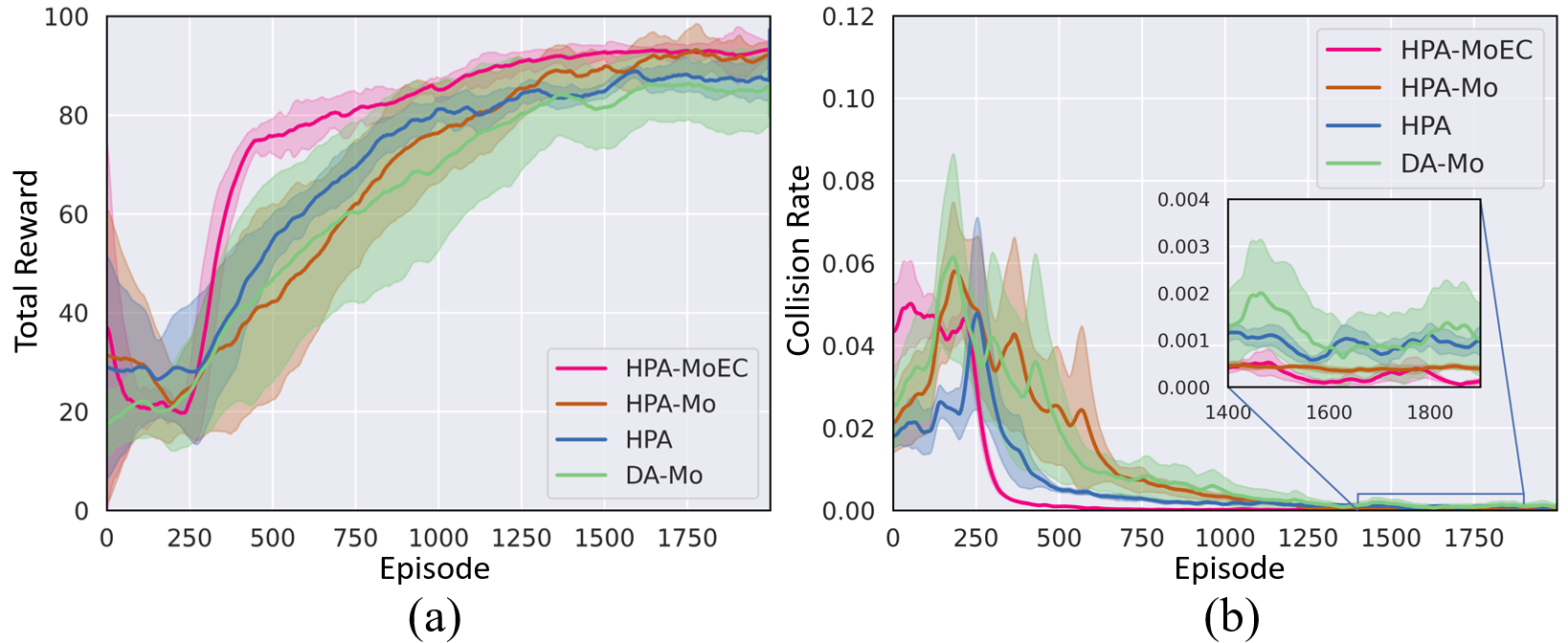}
  \caption{The training process of our framework with ablation baselines quantified by: a) Total Reward and b) Collision Rate.}
  \label{fig:train abla}
\end{figure}

The changes in total reward and collision rate for all ablation baselines during training are shown in Fig.~\ref{fig:train abla}. It is clear that as key components of HPA-MoEC are gradually removed, the performance decreases.

For HPA-Mo, policy convergence is greatly delayed. Compared to HPA-MoEC, HPA-Mo reaches similar final rewards and slightly higher CR. However, its convergence is slower, only reaching around the 1700th episode. In contrast, HPA-MoEC, despite involving more networks, converges around the 1400th episode, suggesting that epistemic uncertainty-based policy exploration improves training efficiency by about 18\%.

For HPA, it shows lower rewards and higher CR at convergence compared to HPA-Mo. This suggests that the designed multi-objective compatible policy evaluation architecture is effective. Utilizing critics that specifically target general driving attributes and safety during policy evaluation can promote driving that is compatible with general performance and safety.

For DA-Mo, the reward it can obtain when converging is the lowest, and the CR is the highest. This shows that hybrid action space structure plays an important role in improving policy execution capabilities. A finer-grained guidance path enhances the correlation between agent output and driving behavior, further improving overall policy performance and safety.

\subsubsection{Testing with Rule-Based SVs}

\begin{figure}[!tb]
  \centering
  \includegraphics[width=0.47\textwidth]{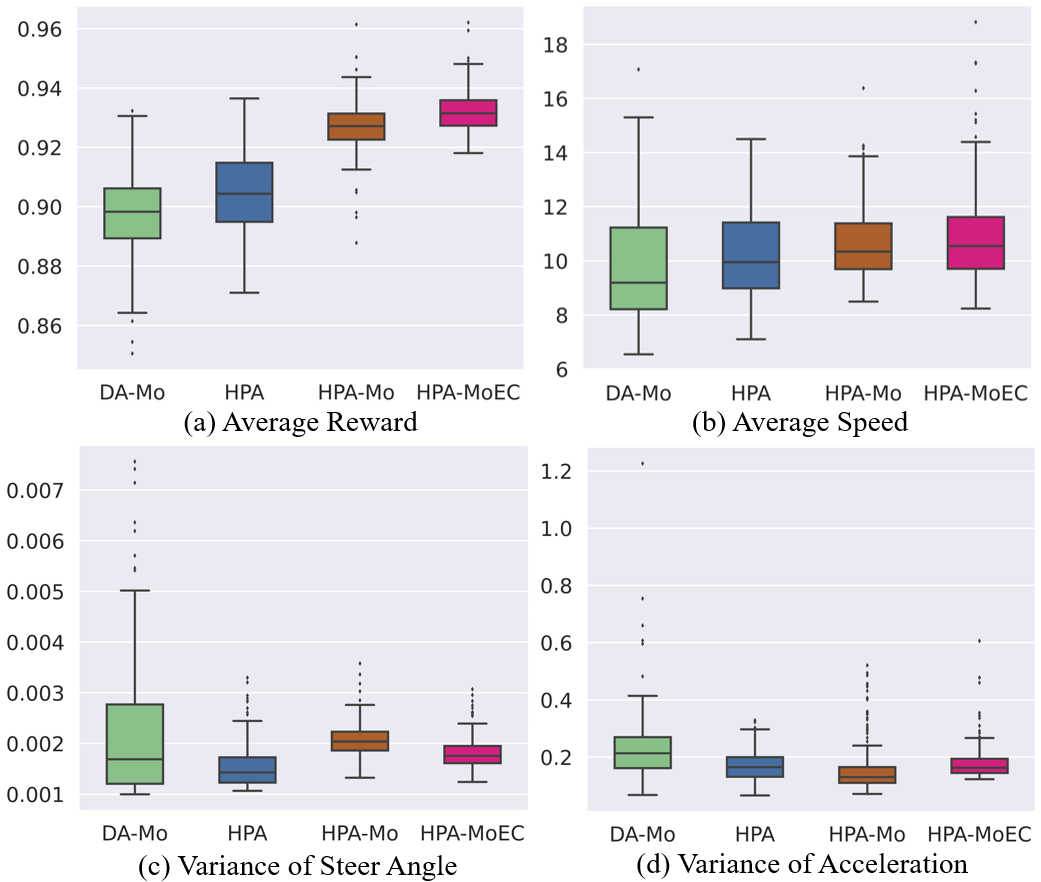}
  \caption{Metrics distribution of ablation study with rule-based SVs.}
  \label{fig:test abla}
\end{figure}

\begin{table}[!tb]  
\caption{Ablative studies for HPA-MoEC with Rule-Based SVs}
\centering  
\begin{threeparttable}
\begin{tabular}{l|cccccc}  
\toprule  
{\textbf{Method}} & \textbf{AR} & \textbf{AS} & \textbf{NL} & \textbf{VS} & \textbf{VA} & \textbf{CR} \\
\midrule  
HPA-MoEC   & 0.932      & 10.87       & 7.14        & 0.0019      & 0.181       & 0.04\%       \\
HPA-Mo     & 0.927      & 10.63       & 6.90        & 0.0020      & 0.160       & 0.03\%       \\
HPA        & 0.905      & 10.36       & 6.11        & 0.0016      & 0.175       & 0.08\%       \\
DA-Mo & 0.898     & 9.22        & 6.43        & 0.0025      & 0.185       & 0.06\%       \\       
\bottomrule  
\end{tabular}
\begin{tablenotes}
\footnotesize
\item[*] All values are averaged over 200 episodes.
\end{tablenotes}
\end{threeparttable}
\label{table:test abla}
\end{table}

The results of the ablation baseline tests, including data distributions and quantitative statistics, are shown in Fig.~\ref{fig:test abla} and Table~\ref{table:test abla}. The HPA-MoEC, with all technology components, demonstrates the best driving performance. As components are progressively removed, the driving performance of the ablation baselines declines accordingly.

HPA-Mo, although slow in policy convergence during training, shows driving performance close to HPA-MoEC in the final testing, with only a slight reduction in AR and AS.

HPA performs worse in both general driving performance and safety, with lower AR and higher CR. Specifically, removing the multi-objective policy evaluation component leads to a significant decrease in AR and, more importantly, nearly a threefold increase in CR for HPA compared to HPA-Mo. This clearly demonstrates that our design maintains the compatibility of the policy with both general performance and safety during testing.

DA-Mo performs the worst across all metrics compared to the other ablation baselines. Notably, removing the hybrid action space results in approximately a 25\% increase in VS compared to HPA, highlighting the larger fluctuations in lateral driving behavior. In addition, its AS decreases by 15\%, with a wider distribution, while the CR increases by 100\%, reflecting a decline in both driving efficiency and safety. Therefore, implementing a hybrid parameterized action space with finer-grained guidance paths helps the agent promote multi-objective driving, particularly in terms of reducing fluctuations in driving behavior.

\subsubsection{Testing in HighD-Dataset}

\begin{figure}[!tb]
  \centering
  \includegraphics[width=0.4675\textwidth]{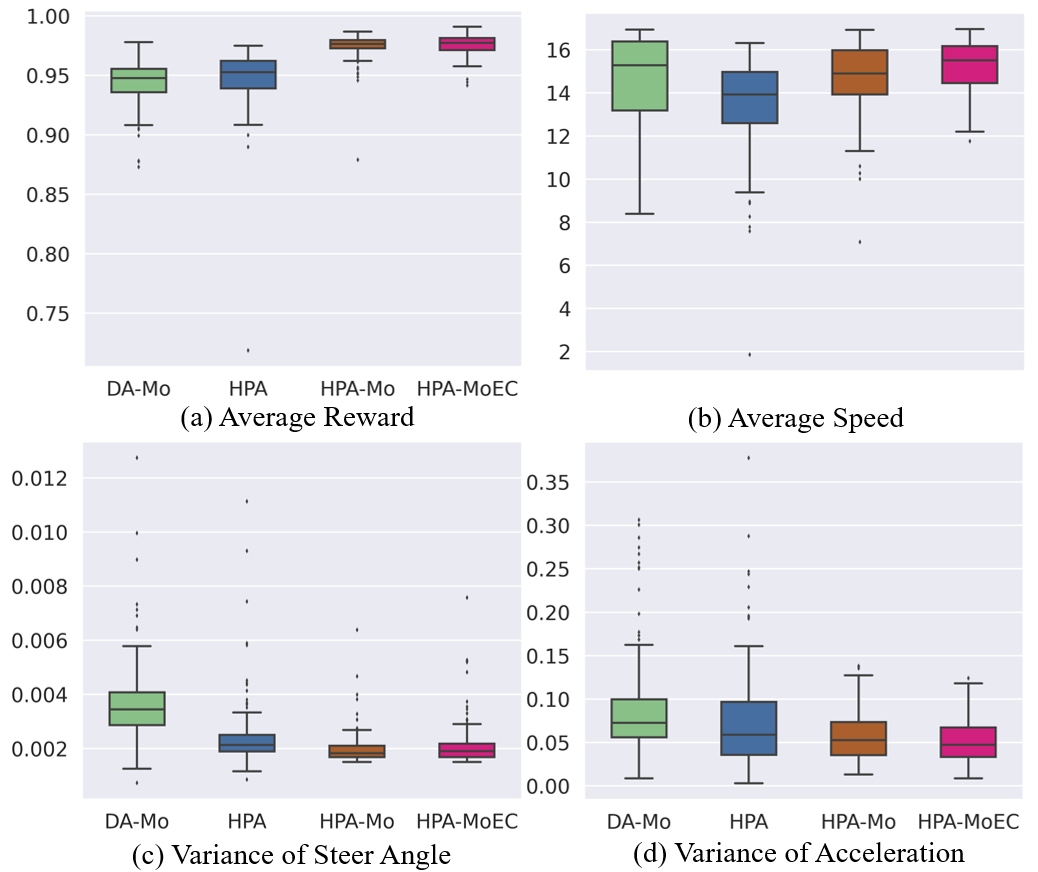}
  \caption{Metrics distribution of ablation study in HighD dataset.}
  \label{fig:highd abla}
\end{figure}

\begin{table}[!tb]  
\caption{Ablative studies for HPA-MoEC in HighD dataset}
\centering  
\begin{threeparttable}
\begin{tabular}{l|cccccc}  
\toprule  
\textbf{Method} & \textbf{AR} & \textbf{AS} & \textbf{NL} & \textbf{VS} & \textbf{VA} & \textbf{CR} \\
\midrule  
HPA-MoEC        & 0.976      & 15.27       & 7.03        & 0.0021      & 0.051       & 0.01\%       \\
HPA-Mo          & 0.975      & 14.71       & 6.88        & 0.0019      & 0.057       & 0.01\%       \\
HPA             & 0.948      & 13.49       & 5.95        & 0.0024      & 0.073       & 0.04\%       \\
DA-Mo      & 0.947      & 14.68       & 6.53        & 0.0036      & 0.076       & 0.04\%       \\       
\bottomrule  
\end{tabular}
\begin{tablenotes}
\footnotesize
\item[*] All values are averaged over 200 episodes.
\end{tablenotes}
\end{threeparttable}
\label{table:highd abla}
\end{table}

The testing results for all ablation baselines in the HighD dataset are shown in Fig.~\ref{fig:highd abla} and Table~\ref{table:highd abla}. HPA-Mo falls slightly below HPA-MoEC in driving efficiency, but both have good driving performance. In contrast, HPA lags clearly behind both previous methods in AS and NL and has a higher CR. The DA-Mo is even worse, accompanying a notable increase in VS. This suggests that the multi-objective policy evaluation architecture and the hybrid parameterized action space with guiding paths still promote the compatibility of the objectives of driving efficiency, action consistency and safety in the HighD dataset.

\subsection{Discussion}

In summary, our HPA-MoEC method outperforms all the RL comparison baselines, where all three key technology components play a significant role in facilitating the learning of a multi-objective compatible policy. The hybrid parameterized action enhances the connection between agent actions and driving behavior by simultaneously outputting finer-grained guiding paths as well as direct acceleration commands. This action space structure promotes multi-objective compatibility, particularly enhancing action consistency by reducing driving behavior fluctuations while maintaining flexibility. The multi-objective policy evaluation architecture guides the agent in improving policy learning by treating general and safety attributes as distinct objectives and building the corresponding reward function and critic. This policy evaluation architecture improves both the driving general performance and safety, demonstrating its ability to achieve multi-objective compatible driving. In addition, the epistemic uncertainty-based policy exploration mechanism accelerates the convergence of multi-objective compatible viable policies, improving the training efficiency. It is also noteworthy that SVs in the HighD dataset exhibit human driving behaviors, which differ significantly from those in simulation traffic. Although HPA-MoEC is trained in simulated traffic, it still achieves strong driving performance when confronted with unfamiliar SVs. This demonstrates that HPA-MoEC possesses strong generalization capabilities and can effectively adapt to unfamiliar environments.

\begin{figure}[!tb]
  \centering
  \includegraphics[width=0.47\textwidth]{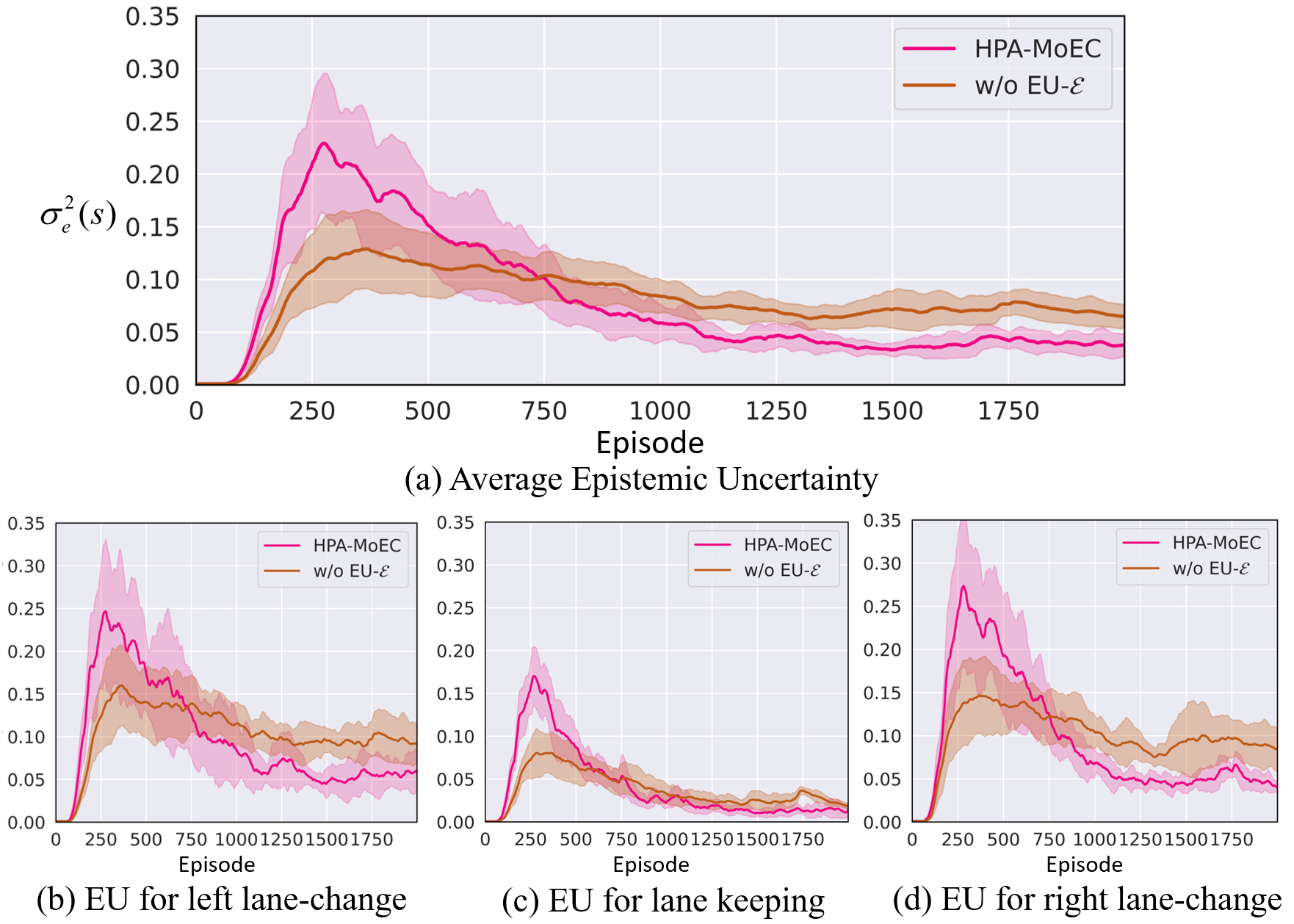}
  \caption{Changes in epistemic uncertainty during training, including: a) average uncertainty, (b) uncertainty for left lane-change, (c) uncertainty for lane keeping, (d) uncertainty for right lane-change.}
  \label{fig:EU}
\end{figure}

Additionally, to better observe the impact of our exploration mechanism on epistemic uncertainty, we introduce the baseline \mbox{'w/o EU-$\cal E$'}, where EU denotes epistemic uncertainty. In this baseline, ensemble-critics generate epistemic uncertainty but do not use it for exploration, instead performing random exploration. The curves in Fig.~\ref{fig:EU} show how epistemic uncertainty evolves throughout the policy improvement process. Our HPA-MoEC experiences higher average uncertainty in the early training phases, and then makes the uncertainty lower more rapidly during exploration. This suggests that HPA-MoEC explores more fully while converging the policy faster than randomized exploration. Further, the changes in epistemic uncertainty for the three lane-change decisions follow a similar trend. Notably, changing lanes—whether to the left or right—results in higher uncertainty compared to lane keeping, suggesting that lane changes involve greater unknowns and risks.

%% file: VI_Conclusion.tex
\section{Conclusion and Future Work}\label{sec:Conclusion}

This paper proposes a Multi-objective Ensemble-Critic (HPA-MoEC) reinforcement learning method with Hybrid Parameterized Action space, capable of efficiently learning multi-objective compatible driving policies. HPA-MoEC adopts a more advanced MORL architecture, in which multiple reward functions guide ensemble-critics to focus on specific driving objectives. Meanwhile, the architecture integrates a hybrid parameterized action space structure, which can simultaneously generate abstract guidance and specific control commands that fit the hybrid road modality. In addition, an uncertainty-based exploration mechanism is developed to achieve faster learning of multi-objective compatible policies. We conduct the training and testing of the policy in both simulated traffic environments and the HighD dataset. The results show that HPA-MoEC effectively learns a multi-objective compatible autonomous driving policy in terms of efficiency, action consistency, and safety. The ablation study further demonstrated the role of technology components in HPA-MoEC in promoting multi-objective compatibility.

One limitation of our study is that the driving scenarios for training and testing are restricted to multi-lane highways. Although this typical structured road environment differs from other road types such as ramps and intersections, the driving objectives of EV in these various scenarios are generally similar: selecting appropriate behavioral goals and interacting with other vehicles. The key difference lies in how the state space is designed to enable the RL agent to comprehensively perceive the environment. Therefore, in future work, we aim to use higher-dimensional perception information (such as BEV images) as the state space to extend the application of HPA-MoEC to more complex traffic scenarios.